\documentclass[10pt]{article} %
\usepackage[accepted]{tmlr}

\usepackage{amsmath,amsfonts,bm}

\def\eqref#1{equation~\ref{#1}}

\def\1{\bm{1}}

\DeclareMathAlphabet{\mathsfit}{\encodingdefault}{\sfdefault}{m}{sl}
\SetMathAlphabet{\mathsfit}{bold}{\encodingdefault}{\sfdefault}{bx}{n}

\usepackage{hyperref}
\hypersetup{
    pdfborder={0 0 0}, %
    colorlinks=true,   %
    linkcolor=blue,    %
    urlcolor=blue,     %
    citecolor=blue     %
}
\usepackage{url}
\usepackage[capitalise]{cleveref}

\newcommand{\loss}{\mathcal{L}}
\newcommand{\trainloss}{\mathcal{L}_{tr}}
\newcommand{\valloss}{\mathcal{L}_{val}}
\newcommand{\mset}{\mathcal{M}}
\newcommand{\dset}{\mathcal{V}}
\newcommand{\trainset}{\mathcal{D}_{tr}}
\newcommand{\valset}{\mathcal{D}_{val}}

\usepackage{graphicx} 
\usepackage{makecell} %
\usepackage{longtable} %
\usepackage{amsthm}
\usepackage{wrapfig}
\usepackage{booktabs}
\usepackage[linesnumbered, ruled,vlined]{algorithm2e}
\usepackage{subfigure}
\usepackage{url}
\usepackage{multirow}

\crefname{algocf}{Algorithm}{Algorithms}  %
\Crefname{algocf}{Algorithm}{Algorithms}  %

\title{BiDoRA: Bi-level Optimization-Based Weight-Decomposed Low-Rank Adaptation}

\author{\name Peijia Qin \email pqin@ucsd.edu \\
      \addr University of California, San Diego
      \AND
      \name Ruiyi Zhang \email ruz048@ucsd.edu \\
      \addr University of California, San Diego
      \AND
      \name Pengtao Xie \email p1xie@ucsd.edu\\
      \addr University of California, San Diego
}

\def\month{07}  %
\def\year{2025} %
\def\openreview{\url{https://openreview.net/forum?id=v2xCm3VYl4}} %

\begin{document}

\maketitle

\begin{abstract}
Parameter-efficient fine-tuning (PEFT) is a flexible and efficient method for adapting large language models (LLMs) to downstream tasks.
Among these methods, weight-decomposed low-rank adaptation (DoRA) is a promising approach that decomposes weight matrices into magnitude and direction components to mimic full fine-tuning (FT) better.
However, DoRA's simultaneous optimization of these components makes it over-expressive, increases the \textit{risk of overfitting}, and creates a \textit{coupled updating pattern} that limits its learning capacity.
To address these issues, we propose \textbf{Bi}-level Optimization-Based Weight-\textbf{D}ecomposed
L\textbf{o}w-\textbf{R}ank \textbf{A}daptation (\textbf{BiDoRA}), a novel PEFT method based on a \textit{bi-level optimization framework}.
BiDoRA fundamentally differs from DoRA by optimizing the magnitude and direction in two separate, asynchronous loops using distinct training and validation data splits.
This decoupled optimization process effectively mitigates overfitting and allows for more flexible updates that align even more closely with FT.
For instance, weight decomposition analysis shows BiDoRA achieves a magnitude-direction update correlation of $\mathbf{-8.042}$, significantly closer to the FT ideal compared to $\mathbf{-1.784}$ for DoRA.
Evaluation of BiDoRA on diverse tasks spanning natural language understanding, generation, token classification, and extremely small biomedical datasets reveals that it consistently outperforms DoRA and a wide range of leading PEFT methods.
This improvement is statistically significant, as demonstrated on the GLUE benchmark where BiDoRA surpasses DoRA with a p-value of $\mathbf{2.4\times10^{-4}}$ in terms of the Wilcoxon signed-rank test. The code for BiDoRA is available at \url{https://github.com/t2ance/BiDoRA}.

\end{abstract}

\section{Introduction}
\begin{figure*}
    \centering
    \includegraphics[width=\linewidth]{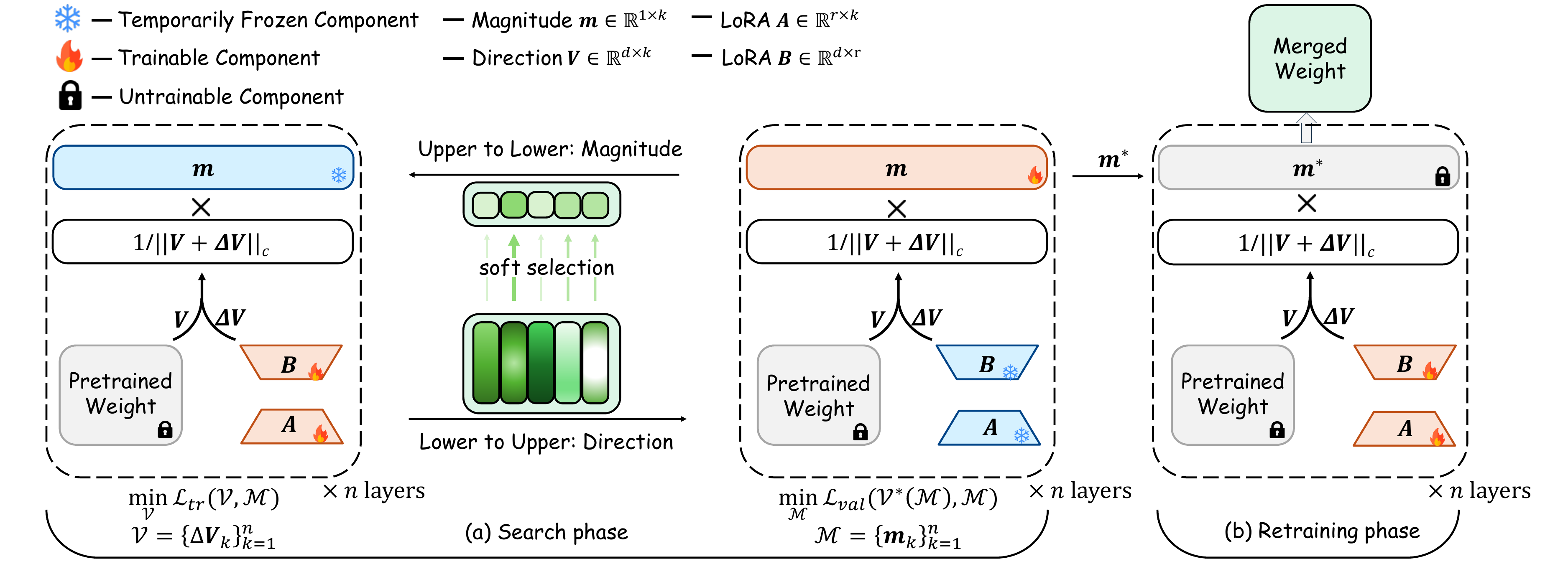}
    \caption{\textbf{An overview of BiDoRA.}
    BiDoRA performs PEFT using a BLO framework.
    \textbf{At the lower level}, BiDoRA learns the direction component $\Delta \mathbf{V}$ of the update matrices using the training split of the downstream dataset.
    \textbf{At the upper level}, BiDoRA optimizes the magnitude component $\mathbf{m}$ with optimized $\Delta \mathbf{V}$ from the lower level, using the validation split of the dataset.
    After determining the optimal magnitude, the direction component undergoes further fine-tuning on a combined set of both training and validation splits to maximize overall performance.
    }
    \label{fig:chart}
\end{figure*}

Large language models (LLMs) \citep{radford2019language,brown2020language} have achieved state-of-the-art results across a broad range of NLP tasks, from natural language understanding (NLU) \citep{wang2019glue} to natural language generation (NLG) \citep{novikova2017e2e}.
Parameter-efficient fine-tuning (PEFT) methods \citep{houlsby2019parameter,hulora} have been introduced as a promising solution for adapting LLMs for downstream data.
PEFT approaches update only a subset of the pre-trained parameters, achieving performance comparable to full-finetuning (FT) while requiring significantly fewer computational resources.

One popular type of PEFT is low-rank adaptation (LoRA, \citet{hulora}), which attaches low-rank matrices to the pre-trained weights and updates only these matrices during fine-tuning.
\citet{liu2024dora} shows that when decomposing the weights into magnitude and direction, their correlation (See \cref{sec:decomposition}) tends to be positive in LoRA, whereas it is negative in FT.
To bridge the training pattern distinction, they introduce an explicit reparameterization of the pre-trained weights matrix.
The method, named DoRA, decomposes the weights into the column-wise product of magnitude and direction, which determines the direction and magnitude of the weight update, respectively.
This approach enables DoRA to share similar learning patterns with FT, thereby outperforming LoRA in multiple tasks.
Nonetheless, DoRA introduces \textbf{additional parameters} and \textbf{over-expressive architecture} compared to LoRA, which can exacerbate overfitting issues when adapting to small downstream datasets (See \cref{tab:gap}).
Furthermore, in DoRA, the magnitude and direction components are optimized concurrently, leading to a \textbf{constrained updating pattern} due to shared optimization setup (e.g., learning rate, optimizer, batch size).

To address the challenges above, we propose BiDoRA, a \textbf{Bi}-level Optimization-Based Weight-\textbf{D}ecomposed
L\textbf{o}w-\textbf{R}ank \textbf{A}daptation method for PEFT.
BiDoRA facilitates an even more flexible updating pattern and mitigates overfitting by separately optimizing the two components on different data splits with distinct optimization levels.
BiDoRA is based on a bi-level optimization (BLO) framework:
At the lower level, the low-rank direction component is updated using the training split, while the magnitude component remains fixed.
At the upper level, the magnitude component is updated by minimizing the loss on the validation split via hypergradient descent.
Subsequently, the direction component is further fine-tuned with the optimal magnitude frozen to maximize the performance.
These two optimization steps are performed iteratively until convergence.
\cref{fig:chart} provides an overview of BiDoRA.

A similar strategy of combating overfitting based on BLO has been utilized in the well-established practice of differentiable neural architecture search (DARTS, \citet{liu2018darts}), where architecture and sub-networks are learned using different dataset splits.
Optimizing the selection variables and sub-networks in a single loop can result in an over-expressive network since the selection variables tend to select all sub-networks to achieve the best expressiveness, which, however, incurs severe overfitting.
In contrast, training the sub-networks with the selection module fixed on the training split while validating the effectiveness of the selection module on the unseen validation split effectively eliminates the risk of overfitting.
Similarly, we treat the \textbf{magnitude component as the architecture} and the \textbf{direction component as the sub-networks} and train these components on \textbf{separate datasets}.
As shown in \cref{tab:gap}, BiDoRA demonstrates better resistance to overfitting compared to DoRA, given the smaller performance gap between the training set and test set.
Furthermore, the asynchronous gradient update steps at the two optimization levels in BiDoRA facilitate better decoupling of the two components, leading to a more flexible update pattern that closely resembles FT.
As illustrated in \cref{fig:wdaquery}, the updates across different layers using BiDoRA have a correlation value that is closest to that of FT, highlighting its superior learning capability compared to both DoRA and LoRA.

Our work makes the following key contributions:
\begin{itemize}
    \item We propose \textbf{BiDoRA}, a novel PEFT method based on bi-level optimization.
    In contrast to DoRA, which trains the magnitude and direction components on a single dataset, BiDoRA optimizes these components at different optimization levels.
    \item Our strategy effectively mitigates the risk of overfitting and results in a parameter update pattern that more closely resembles full fine-tuning.
    \item Extensive experiments on various downstream tasks highlight the superior performance of BiDoRA.
    BiDoRA consistently surpasses several baseline methods, including LoRA and DoRA.
\end{itemize}

\section{Related Work}
\subsection{Parameter Efficient Fine-Tuning Methods}
Parameter-efficient fine-tuning (PEFT) methods aim to reduce the high costs associated with full fine-tuning large-scale models by updating only a relatively small subset of pre-trained parameters, rather than the entire model, to adapt to downstream tasks.
Existing PEFT methods can be mainly categorized into three types.

\textbf{The first category, known as adapter-based methods,} injects additional trainable modules into the original frozen backbone.
For instance, \citet{houlsby2019parameter} suggests adding linear modules in sequence to existing layers, while \citet{he2021towards} proposes integrating these modules in parallel with the original layers to enhance performance.
Recent advances include SAN \citep{xu2023side}, FADA \citep{bi2024learning}, and SET \citep{yi2024learning}.
SAN presents a side adapter network attached to a frozen CLIP model, which contains two branches for predicting mask proposals and attention biases.
FADA introduces a frequency-adapted learning scheme that uses the Haar wavelet transform to decompose frozen features into low- and high-frequency components, which are processed separately to enhance domain generalization.
SET proposes a spectral-decomposed token learning framework that leverages the Fast Fourier Transform to separate frozen features into amplitude and phase components, enhancing them with spectral tokens and attention optimization.

\textbf{The second category is prompt tuning methods}, which add extra soft tokens (prompts) to the initial input.
During the fine-tuning stage, only these trainable soft tokens are updated, as demonstrated in works such as \citet{lester2021power} and \citet{razdaibiedina2023residual}.
Unfortunately, the first two categories lead to increased inference latency compared to fully fine-tuned models.

\textbf{The third prominent category focuses on low-rank adaptation}, pioneered by LoRA \citep{hu2022lora}.
LoRA injects trainable, low-rank matrices into a model's layers, freezing the original weights.
A key advantage is that these low-rank updates can be merged into the original weights before inference, thus incurring no additional latency.
Subsequent works have aimed to improve LoRA's efficiency and performance.
For instance, AdaLoRA \citep{zhang2023adaptive} dynamically reallocates the parameter budget based on the importance scores of weight matrices.
\citet{zhang-etal-2024-autolora} uses meta-learning to search for the optimal rank of LoRA matrices, further improving its performance on downstream tasks.
Pushing parameter efficiency further, VeRA \citep{kopiczko2023vera} employs a single pair of shared low-rank matrices across all layers, while AFLoRA \citep{liu2024aflora} freezes a portion of adaptation parameters based on a learned score.
A distinct sub-direction has emerged that performs adaptation in the frequency domain, including FourierFT \citep{gao2024parameter}, LaMDA \citep{azizi2024lamda}, SSH \citep{shen2025ssh}, and MaCP \citep{shen2025macp}.
These methods learn updates in transformed spectral spaces, such as the Fourier, discrete Hartley, or discrete cosine domains, rather than directly in the weight space.
Other research has focused on bridging the performance gap between LoRA and full fine-tuning.
\citet{liu2024dora} found that LoRA's update patterns differ significantly from full fine-tuning, potentially constraining its learning capacity. To mitigate this, they proposed DoRA \citep{liu2024dora}, which decomposes pre-trained weights into magnitude and direction components and uses LoRA for efficient directional updates, better mimicking full fine-tuning.

\subsection{Bi-level Optimization}
Bi-level optimization (BLO) has been widely applied in various machine learning tasks, including meta-learning \citep{finn2017model,rajeswaran2019meta}, neural architecture search (NAS) \citep{liu2018darts,zhang2021idarts}, and hyperparameter optimization \citep{lorraine2020optimizing,franceschi2017forward}.
Despite its wide usage, solving BLO problems can be challenging due to the inherent nature of nested optimization problems.
Several algorithms have been proposed to address this challenge, including zeroth-order methods such as Bayesian optimization \citep{cui2019new} and first-order algorithms based on hypergradients \citep{pearlmutter2008reverse,lorraine2020optimizing}.
Among these approaches, gradient-based BLO has received significant attention because it can scale to high-dimensional problems with a large number of trainable parameters.

Inspired by NAS, where a bi-level approach is used to learn an architecture and its sub-network weights on separate data splits to prevent overfitting, we adapt the BLO framework to parameter-efficient fine-tuning (PEFT), specifically for the weight-decomposed adaptation introduced by DoRA.
Unlike in NAS, where BLO searches for a network architecture, BiDoRA repurposes it to decouple the optimization of a weight matrix's two components: magnitude and direction. This approach marks a significant departure from previous PEFT methods like LoRA and DoRA, which optimize all trainable parameters simultaneously on a single dataset.
In this work, we extend the application of gradient-based BLO to develop a robust and effective PEFT method for pre-trained models.
By assigning the magnitude and direction components to different optimization levels with distinct data splits, BiDoRA creates a decoupled, flexible updating pattern that better mitigates overfitting and more closely resembles the learning behavior of full fine-tuning.

\section{Preliminary}
LoRA \citep{hulora} involves attaching the product of two low-rank matrices to the pre-trained weights and fine-tuning these low-rank matrices on downstream datasets with the pre-trained weights frozen.
It is based on the assumption that parameter updates made during fine-tuning exhibit a low intrinsic rank.
Formally, given a pre-trained weight matrix $\mathbf{W_0} \in \mathbb{R}^{d \times k}$, LoRA attaches a low-rank update matrix $\Delta \mathbf{W} \in \mathbb{R}^{d \times k}$ to the pre-trained weight. This update matrix can be decomposed as $\Delta \mathbf{W} = \mathbf{B}\mathbf{A}$, where $\mathbf{B} \in \mathbb{R}^{d \times r}$ and $\mathbf{A} \in \mathbb{R}^{r \times k}$ are two low-rank matrices, with $r \ll \min(d, k)$.
Consequently, the weight matrix $\mathbf{W}^{\prime}$ is represented as follows:
\begin{equation}
    \mathbf{W}^{\prime}=\mathbf{W_0}+\Delta \mathbf{W}=\mathbf{W_0}+\mathbf{B}\mathbf{A}
\end{equation}
In this setup, only the LoRA matrix $\Delta \mathbf{W}$ is updated.
\citet{liu2024dora} found that LoRA and full fine-tuning exhibit different learning patterns by performing weight decomposition on fine-tuned weight matrices (See \cref{sec:decomposition}).
To bridge this discrepancy, weight-decomposed low-rank adaptation (DoRA, \citet{liu2024dora}) further reparameterizes the weight matrices by explicitly decomposing them into learnable magnitude and direction components. Formally, DoRA performs adaptation as follows:
\begin{equation}
        \mathbf{W}^{\prime}=\mathbf{m}\frac{\mathbf{V}+\Delta \mathbf{V}}{\|\mathbf{V}+\Delta \mathbf{V}\|_c}=
    \mathbf{m}\frac{\mathbf{W_0}+\mathbf{B}\mathbf{A}}{\|\mathbf{W_0}+\mathbf{B}\mathbf{A}\|_c}
    \label{eq:dora}
\end{equation}
where $\Delta \mathbf{V}$ is a product of two learnable low-rank matrices, $\mathbf{B}$ and $\mathbf{A}$, while the magnitude component $\mathbf{m} \in \mathbb{R}^{1\times k}$ is a learnable vector.
Here, $\|\cdot\|_c$ represents the vector-wise norm of a matrix computed across each column, using the $L_2$ norm.
In DoRA, both components are optimized concurrently on a single downstream dataset.
In this work, we aim to improve DoRA by further decoupling the training of the two components.

\section{Methods}

\subsection{Overview of BiDoRA}
Our method, BiDoRA, optimizes the trainable parameters in DoRA layers by solving a BLO problem.
Let $\mathcal{M} = \{\mathbf{m}_1, \mathbf{m}_2, \ldots, \mathbf{m}_n\}$ denote the set of magnitude components for all $n$ DoRA modules, and $\mathcal{V} = \{\Delta \mathbf{V}_1, \Delta \mathbf{V}_2, \ldots, \Delta \mathbf{V}_n\}$ denote the set of corresponding direction components.
Specifically, we first learn the direction components $\mathcal{V}^*(\mathcal{M})$ on the training split of the downstream dataset $\mathcal{D}_{tr}$ at the lower level.
The magnitude component $\mathcal{M}$ is tentatively fixed at this level; thus, the resulting optimal direction component $\mathcal{V}^*(\mathcal{M})$ is a function of $\mathcal{M}$. At the upper level, we determine the optimal magnitude component $\mathcal{M}^*$ by optimizing the loss on a validation split $\mathcal{D}_{val}$. In practice, $\mathcal{D}_{tr}$ and $\mathcal{D}_{val}$ are typically created by splitting the original training set without using additional data.
This BLO problem is solved using an efficient gradient-based algorithm, where parameters at two levels are optimized iteratively until convergence.
While this work focuses on the empirical validation of BiDoRA, our choice of optimization strategy is grounded in established theoretical research.
The convergence properties of similar gradient-based bi-level algorithms have been previously analyzed \citep{pedregosa2016hyperparameter,rajeswaran2019meta}, providing confidence in the stability of our training procedure. 
Furthermore, the ability of such frameworks to improve generalization—a core objective of BiDoRA—has also been formally studied \citep{bao2021stability}, supporting the rationale that our approach can mitigate overfitting.

\subsection{Orthogonal Regularization}
\label{sec:or}
A central goal of BiDoRA is to learn the two disentangled components of a weight update: magnitude and direction.
The direction component, $\Delta \mathbf{V}$, is responsible for finding a low-rank basis for the update directions.
To maximize the expressive power of this component and prevent overfitting, its basis vectors (i.e., the columns of the direction matrix) should be as diverse and non-redundant as possible.

The orthogonality of neural network weights has been identified as a beneficial property~\citep{Bansal2018CanWG} and can effectively mitigate the overfitting issue~\citep{pmlr-v80-balestriero18b}.
By enforcing orthogonality, the direction vectors are constrained to represent distinct, independent pathways for updates. This ensures that the limited parameter budget of the low-rank matrix is used efficiently to explore the solution space.
Therefore, we define a Gram regularization loss~\citep{Xie2017gram} for the direction component:
\newcommand{\norm}[1]{\left\lVert#1\right\rVert}
\begin{equation}
    \mathcal{R}(\dset)=\sum_{k=1}^n\norm{(\mathbf{V}_k+\Delta \mathbf{V}_k)^\top(\mathbf{V}_k+\Delta \mathbf{V}_k)-\mathbf{I}}_F^2
    \label{eq:regularizer}
\end{equation}
where $\mathbf{I}$ is the identity matrix and $\|\cdot\|_F$ denotes the Frobenius norm. 
Intuitively, $\mathcal{R}(\dset)$ encourages each column of the direction matrix, representing a specific direction, to be orthogonal to one another. 
Since each column has already been normalized (equivalent to projected to the unit sphere), this also prompts each column to be far away from the other, thereby reducing the redundancy of parameters.
The effectiveness of this constraint is empirically validated in our ablation study (See \cref{tab:ablation}), which shows a consistent performance improvement resulting from the enhanced generalization ability of the learned direction component.

\subsection{A Bi-level Optimization Framework}
\paragraph{Lower level.}
At the lower level, we train the low-rank direction component $\mathcal{V}$ by minimizing a loss $\trainloss$ defined on the training set $\trainset$.
The overall training objective at this level is $\trainloss(\dset,\mset) = \loss(\dset, \mset; \trainset) + \gamma \mathcal{R}(\dset)$. Here, $\loss$ represents the fine-tuning loss, given the low-rank direction component $\dset$, the magnitude component $\mset$, and the training split $\trainset$ of the downstream dataset.
$\mathcal{R}(\dset)$ is the orthogonal regularizer defined in \cref{eq:regularizer}, with $\gamma$ as a trade-off hyperparameter.
In this level, we only update $\dset$ while keeping $\mset$ fixed, resulting in the following optimization problem:
\begin{align}
\dset^*(\mset)=\arg\min_{\dset}\:\trainloss(\dset,\mset)  \label{eq:lower}
\end{align}
where $\dset^{*}(\mset)$ denotes the optimal solution for $\dset$ in this problem, which is a function of $\mset$.
\paragraph{Upper level.}
At the upper level, we validate the previously fixed magnitudes $\mset$ on the validation set $\valset$, using the optimal direction component $\dset^*(\mset)$ that was learned at the lower level.
This results in a validation loss $\valloss(\dset^*(\mset),\mset) = \loss(\dset^*(\mset), \mset; \valset)$.
We determine the optimal magnitude component $\mset$ by minimizing this validation loss:
\begin{align}
  \min_{\mset} \valloss(\dset^*(\mset),\mset)
  \label{eq:upper}
\end{align}
\paragraph{A bi-level optimization framework.}
Integrating the two levels of optimization problems, we have the following BLO framework:

\begin{align}
\label{eq:bilevel}
&\min_{\mset}\:\valloss(\dset^*(\mset),\mset)\notag  \\
s.t.\quad&\:\dset^*(\mset)=\arg\min_{\dset}\:\trainloss(\dset,\mset)
\end{align}

Note that these two levels of optimization problems are mutually dependent on each other.
The solution of the optimization problem at the lower level, $\dset^*(\mset)$, serves as a parameter for the upper-level problem, while the optimization variable $\mset$ at the upper level acts as a parameter for the lower-level problem.
By solving these two interconnected problems jointly, we can learn the optimal magnitude component $\mset^*$ and incremental direction matrices $\dset^*$ in an end-to-end manner.

Two reasons exist behind the choice of setting the magnitude component as the upper level instead of the converse one: 1) In literature, the upper level usually has fewer parameters than the lower level.
In our case, the design of setting the magnitude of complexity $\mathcal{O}(k)$ as the upper level and the direction of complexity $\mathcal{O}(dr+kr)$ as the lower level is consistent with the common practice.
2) BiDoRA resembles the DARTS method \citep{liu2018darts} in NAS literature, where the subnets are selected by a selection variable.
Specifically, the magnitude vector resembles a selection variable on the direction matrix by softly selecting each direction (subnets) via scaling.
\paragraph{Optimization algorithm.}

\begin{algorithm}[t!]
\caption{BiDoRA}
\label{alg:bidora}
\KwIn{Training dataset $\trainset$ and validation dataset $\valset$}

Initialize trainable magnitude components $\mset=\{\mathbf{m}_k\}_{k=1}^n$ and low-rank direction components $\dset=\{\Delta \mathbf{V}_k\}_{k=1}^n=\{\{\mathbf{A}_k\}_{k=1}^n, \{\mathbf{B}_k\}_{k=1}^n\}$  

\textcolor{orange}{// \underline{Search Phase}}

\While{\textcolor{orange}{not converged}}{
\textcolor{orange}{
Update magnitude $\mset$ by descending $\nabla_{\mset}{\valloss (\dset-\xi\nabla_{\dset}{\trainloss(\dset, \mset)}, \mset) }$
}

\textcolor{orange}{
Update direction $\dset$ by descending $\nabla_{\dset}{\trainloss(\dset, \mset)}$
}
}
\textcolor{orange}{Derive the optimal magnitude $\mset^*=\{m^*_k\}_{k=1}^n$}

\textcolor{violet}{// \underline{Retraining Phase}}

\textcolor{violet}{Train $\dset$ until converge using $\trainset \bigcup \valset$ and derive the optimal direction $\dset^*$}

\KwOut{$\dset^*$ and $\mset^*$}

\end{algorithm}

We use a gradient-based optimization algorithm \citep{choebetty} to solve the BLO problem presented in \cref{eq:bilevel}.
A significant challenge in this process is the computation of the upper-level loss gradient with respect to the magnitude component $\mset$, as this gradient depends on the optimal solution of the lower-level problem, $\dset^*(\mset)$.
For deep neural networks, the lower-level objective is non-convex, meaning that finding the true optimal solution $\dset^*(\mset)$ would require running its optimization process to full convergence.
Performing this complete inner optimization for every single update of the upper-level variable $\mset$ is computationally intractable.

To address this issue, we use the following one-step-unrolled approximation of $\dset^*(\mset)$ inspired by previous work~\citep{liu2018darts}: 

\begin{align}
    \nabla_{\mset}{\valloss (\dset^*(\mset), \mset) }
    \approx \nabla_{\mset}{\valloss (\dset-\xi\nabla_{\dset}{\trainloss(\dset, \mset)}, \mset) } \label{eq:firstorder}
\end{align}

where $\xi$ is the learning rate at the lower level, and the one-step-unrolled model $\bar{\dset} = \dset - \xi\nabla_{\dset}{\trainloss(\dset, \mset)}$ is used as a surrogate for the optimal solution $\dset^*(\mset)$.
We then compute the approximated gradient as follows:
{
\begin{align}
    &\nabla_{\mset}{\valloss (\dset-\xi\nabla_{\dset}{\trainloss(\dset, \mset)}, \mset) } \notag
    \\
    = & \nabla_{\mset}\valloss (\bar{\dset}, \mset)-\xi \nabla^2_{\mset,\dset}\trainloss(\dset, \mset)\nabla_{\bar{\dset}}\valloss(\bar{\dset}, \mset) \label{eq:secondorder}
    \\ 
    \approx & \nabla_{\mset}\valloss (\bar{\dset}, \mset) - \xi \frac{\nabla_{\mset} \trainloss(\dset^+, \mset) - \nabla_{\mset} \trainloss (\dset^-, \mset) }{2\epsilon} \label{eq:darts}
\end{align}
}

where $\epsilon$ is a small scalar and $\dset^\pm = \dset \pm \epsilon\nabla_{\bar{\dset}}{\valloss (\bar{\dset}, \mset)}$. Since directly computing the matrix-vector multiplication term in \cref{eq:secondorder} is computationally expensive, we use finite difference to approximate this product as in \cref{eq:darts}, following \citet{liu2018darts}.
As detailed in \cref{alg:bidora}, the direction component $\dset$ and the magnitude component $\mset$ are updated using gradient descent iteratively until convergence.
After acquiring the optimal magnitudes $\mset^*$ through the process above, the direction component $\dset$ is retrained on the union of training and validation splits to achieve the best performance on downstream tasks, resulting in the final learned $\dset^*$.
All splits are intentionally used during retraining to maximize data utilization and performance.

In practice, the convergence of the search phase is determined by the evaluation metric at the upper level.
For the subsequent retraining phase, we adopt a stopping criterion similar to DoRA's, observing performance on a separate, held-out test set that is not used during training.
\begin{table*}[htbp!]
    \centering
    \caption{RoBERTa\textsubscript{base/large} (R\textsubscript{b/l}) and DeBERTa\textsubscript{XXL} (D\textsubscript{XXL}) with different fine-tuning methods on the GLUE benchmark \citep{wang2019glue}.
    A higher value is better for all datasets.
    The best results are shown in \textbf{bold}.
}
\vspace{0.2cm}
    \setlength{\tabcolsep}{3pt}
    \begin{tabular}{c|c|cccccccccc} \toprule
        Method & \#Params & MNLI & SST-2 & MRPC & CoLA & QNLI & QQP & RTE & STS-B & Avg.  \\ \midrule
        R\textsubscript{b}(FT)     & $125.0$M & $90.3$ & $94.8$ & $89.3$ & $61.6$ & $86.7$& $92.8$& $76.9$& $91.2$& $85.5$ \\ \midrule
        R\textsubscript{b}(Adapter)& $0.9$  M & $86.5$ & $94.0$ & $88.4$ & $58.8$ & $92.5$ &$89.1$ & $71.2$& $89.9$& $83.8$ \\
        R\textsubscript{b}(LoRA)   & $0.15$ M & $86.8$ & $94.3$ & $88.0$ & $60.3$ & $\textbf{93.0}$ & $89.6$ & $72.9$ & $90.1$ & $84.4$ \\ 
        R\textsubscript{b}(DoRA)   & $0.17$ M & $86.8$ & $94.2$ & $89.2$ & $60.5$ & $92.9$ & $89.6$ & $73.2$ & $\textbf{90.2}$ & $84.6$ \\ 
        R\textsubscript{b}(BiDoRA) & $0.17$ M & $\textbf{87.1}$ & $\textbf{94.4}$ & $\textbf{89.4}$ & $\textbf{61.3}$ & $92.7$ & $\textbf{90.6}$ & $\textbf{76.0}$ & $90.1$ & $\textbf{85.2}$\\ \midrule
        R\textsubscript{l}(FT)     & $355.0$M& $90.2$ & $96.4$&$90.9$&$68.0$& $94.7$& $92.2 $&$86.6 $&$92.4 $&$88.9$ \\ \midrule
        R\textsubscript{l}(Adapter)& $0.8$M &$90.3$ &$96.3$ &$87.7$& $66.3$& $\textbf{94.7}$ &$91.5$& $72.9$ &$91.5$& $86.4$ \\ 
        R\textsubscript{l}(LoRA)   & $0.39$ M &$\textbf{90.6}$ & $96.3$ & $90.0$ & $66.9$ & $94.5$ & $91.2$ & $86.3$ & $91.7$ & $88.4$ \\ 
        R\textsubscript{l}(DoRA)   & $0.39$ M & $\textbf{90.6}$ & $\textbf{96.4}$ & $89.8$ & $65.8$ & $\textbf{94.7}$ & $91.2$ & $86.6$ & $\textbf{92.0}$ & $88.4$ \\ 
        R\textsubscript{l}(BiDoRA) & $0.39$ M & $\textbf{90.6}$ & $96.1$ & $\textbf{90.1}$ & $\textbf{67.0}$ & $94.6$ & $\textbf{91.7}$ & $\textbf{86.9}$ & $\textbf{92.0}$ & $\textbf{88.6}$ 
        \\ \midrule
        D\textsubscript{XXL}(DoRA)     & $4.9$M& 91.2 & \textbf{96.3} & 92.3 & 71.1 & \textbf{95.3} & 91.6 & 91.8 & \textbf{90.8} & 90.0\\  
        D\textsubscript{XXL}(BiDoRA)     & $4.9$M& \textbf{91.7} & \textbf{96.3} & \textbf{92.6} & \textbf{72.3} & 95.2 & \textbf{92.0} & \textbf{92.3} & \textbf{90.8} & \textbf{90.4} \\
        \bottomrule
\end{tabular}
    \label{tab:nlu}
\end{table*}

\section{Experiments}
\subsection{Experimental Setup}
We compare BiDoRA with several PEFT methods, including Full Fine-Tuning (FT), Adapter Tuning \citep{houlsby2019parameter,lin2020exploring,ruckle2020adapterdrop,pfeiffer2020adapterfusion}, LoRA \citep{hu2022lora}, A AdaLoRA \citep{zhang2023adaptive}, DoRA \citep{liu2024dora}, VeRA \citep{kopiczko2023vera}, FourierFT \citep{gao2024parameter}, AFLoRA \citep{liu2024aflora}, LaMDA \citep{azizi2024lamda}, SSH \citep{shen2025ssh}, MaCP \citep{shen2025macp}.
BiDoRA \textbf{does not use any additional data} compared to other baselines, as we create the validation set for upper-level optimization by splitting the original training set with an $8$:$2$ ratio for all tasks.
Detailed descriptions of these baseline methods are provided in \cref{sec:baselines}.

Our experiments cover a wide range of tasks, including natural language understanding (\cref{sec:nlu}), extremely small biomedical datasets (\cref{sec:biomed}), natural language generation (\cref{sec:nlg}), and token classification (\cref{sec:token}).
Please refer to the detailed dataset settings and experimental settings in \cref{sec:appendixdatasetsmodels} and \cref{sec:appendixexperimentalsettings}, respectively.
Our implementation is based on the Huggingface Transformers library \citep{wolf2019huggingface} and the Betty library \citep{choebetty}.

\subsection{Experiments on Natural Language Understanding Tasks}
\label{sec:nlu}

In this section, we evaluate the performance of BiDoRA on NLU tasks.

\paragraph{Main results.}

\cref{tab:nlu} presents the results of fine-tuning the RoBERTa-base, RoBERTa-large, and DeBERTa XXL models on the GLUE benchmark with baseline PEFT methods and BiDoRA. The results show that BiDoRA achieves superior or comparable performance compared to baseline methods across all datasets with the same number of trainable parameters.
The superior performance of BiDoRA verifies the effectiveness of its BLO mechanism.
By training the magnitude and direction components on two distinct sub-datasets, BiDoRA enhances the flexibility of the learning process and improves learning capacity compared to DoRA, resulting in a performance boost.

We also present an experiment on the GLUE benchmark with the RoBERTa-base model, on a larger, wide range of baselines, following the settings from \citet{shen2025ssh} and \citet{shen2025macp} and citing their reported baseline results for reference.
The results in \cref{tab:additional} indicate that BiDoRA consistently outperforms all baselines, including DoRA, across these diverse NLU tasks, demonstrating its robust generalization capability.

\begin{table*}[!ht]
\centering
\caption{Performance of various fine-tuning methods on the GLUE benchmark for the RoBERTa-base model. The best ones are highlighted by \textbf{bold} and the second ones are highlighted by \textit{italic}.}
\vspace{0.2cm}
\begin{tabular}{c|ccccccccc}
\toprule
Model  & SST-2 & MRPC & CoLA & QNLI & RTE & STS-B & Avg. \\
\midrule
FT  & 94.8 & 90.2 & 63.6 & 92.8 & 78.7 & \textit{91.2} & 85.22 \\
BitFit & 93.7 & \textbf{92.7} & 62.0 & 91.8 & 81.5 & 90.8 & 85.42 \\
Adpt\textsuperscript{D}  & 94.7 & 88.4 & 62.6 & 93.0 & 75.9 & 90.3 & 84.15 \\
LoRA  & \textit{95.1} & 89.7 & 63.4 & \textit{93.3} & 78.4 & \textbf{91.5} & 85.23 \\
AdaLoRA  & 94.5 & 88.7 & 62.0 & 93.1 & 81.0 & 90.5 & 84.97 \\
AFLoRA & 94.1 & 89.3 & 63.5 & 91.3 & 77.2 & 90.6 & 84.33 \\
LaMDA &  94.6 & 89.7 & 64.9 & 91.7 & 78.2 & 90.4 & 84.92 \\
VeRA & 94.6 & 89.5 & 65.6 & 91.8 & 78.7 & 90.7 & 85.15 \\
FourierFT  & 94.2 & 90.0 & 63.8 & 92.2 & 79.1 & 90.8 & 85.02 \\
SSH  & 94.1 & \textit{91.2} & 63.6 & 92.4 & 80.5 & 90.9 & 85.46 \\
MaCP  & 94.2 & 89.7 & 64.6 & 92.4 & 80.7 & 90.9 & 85.42 \\
\midrule
DoRA ($r=8$)      & 94.9   & 89.9   & 63.7   & \textit{93.3}   & 78.9   & \textbf{91.5}   & 85.37 \\
BiDoRA ($r=8$)    & \textbf{95.7}   & 90.2   & \textit{65.8}   & \textbf{93.4}   & 79.4   & 90.5   & 85.83 \\
DoRA ($r=16$)     & 94.8   & 90.4   & 65.6   & 93.1   & \textit{81.9}   & 90.7   & \textit{86.08} \\
BiDoRA ($r=16$)   & 95.0   & 90.8   & \textbf{66.7}   & \textit{93.3}   & \textbf{82.6}   & 90.9   & \textbf{86.55} \\
\bottomrule
\end{tabular}
\label{tab:additional}
\end{table*}

\begin{table}
    \centering
    \caption{Quantitative performance gap between training and test sets for DoRA and BiDoRA using the RoBERTa-base model. The gap is calculated as the training metric minus the test metric, where a smaller value indicates less overfitting.}
    \vspace{0.2cm}
    \begin{tabular}{c|ccccccc}\toprule
        Method & SST-2 & MRPC & CoLA & QNLI & RTE & STS-B & Avg.   \\ \midrule
        DoRA   &     2.0&9.5&32.5&6.6&18.0&8.8&12.9       \\
        BiDoRA &     \textbf{1.7}&\textbf{7.0}&\textbf{23.3}&\textbf{0.2}&\textbf{14.0}&\textbf{4.7}&\textbf{8.5}      \\
        \bottomrule
    \end{tabular}
    \label{tab:gap}
\end{table}

\paragraph{Robustness of BiDoRA towards different rank settings.}

We explore the impact of different rank configurations on BiDoRA and DoRA, evaluating them with ranks of $8$ and $16$ in addition to the rank of $4$ used in \cref{tab:nlu}.
The average accuracies reported in \cref{tab:additional} demonstrate that BiDoRA consistently surpasses DoRA across all rank configurations, highlighting its resilience and superior performance regardless of the rank setting.

\paragraph{Performance gap between training and testing set.}

\begin{figure*}
    \centering
    \includegraphics[width=\linewidth]{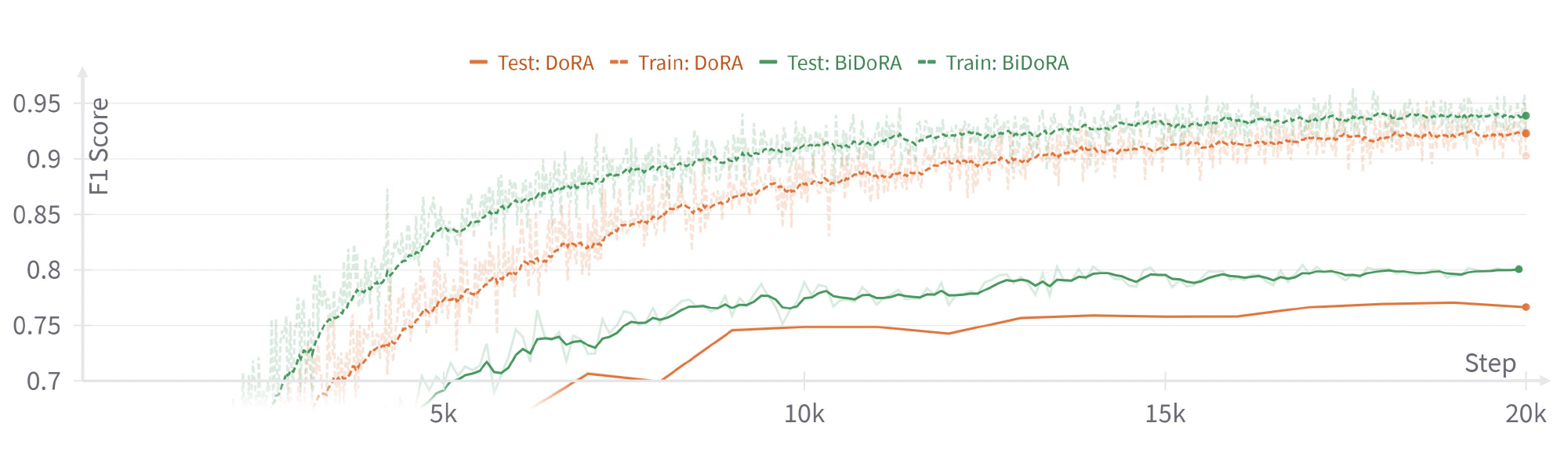}
    \caption{Training and test accuracy versus global training steps on the ModHayes split of the Reuters21578 dataset \citep{padmanabhan2016topic} when fine-tuning a RoBERTa-base model using DoRA and BiDoRA.
    The training and test curves for DoRA show a larger gap compared to BiDoRA, highlighting the effectiveness of our method in reducing overfitting.
    }
    \label{fig:overfitting}
\end{figure*}

As visualized in \cref{fig:overfitting}, BiDoRA achieves a smaller gap between the training and test curves.
Quantitatively, \cref{tab:gap} presents this performance gap on the RoBERTa-base model.
The training set metric is calculated as a moving average of the per-batch metric with a decay ratio of $0.99$.
Since BiDoRA has two training loops, its training metric is a weighted average ($0.8 \times \text{inner-loop-metric} + 0.2 \times \text{outer-loop-metric}$), based on the data split size, $\text{inner}:\text{outer}=8:2$, in our case.
The results show that the performance gap for BiDoRA is consistently lower than that of DoRA across all datasets.
This suggests that DoRA is more prone to overfitting, an issue that BiDoRA effectively addresses.

\subsection{Experiments on Extremely Small Datasets}
\label{sec:biomed}

\begin{table}[htbp]
    \centering
    \caption{
    Fine-tuning ESM on the thermostability prediction task \citep{chenhotprotein} (left), the BBP task \citep{dai2021bbppred} (middle), and the MIC task \citep{ledesma2023engineered} (right).
    A higher value is better for all metrics except for MSE. The best results are highlighted in \textbf{bold}.
    }
    \setlength{\tabcolsep}{1pt}
    \vspace{0.2cm}
    \small
    \begin{minipage}{0.45\textwidth}
        \centering
        \begin{tabular}{c|c|cccc} \toprule
             Methods & \#Params & Accuracy & Precision & Recall & F1 \\ \midrule
             FT & 652.7M & 79.8 & 81.2 & 79.8 & 78.4\\ \midrule
             LoRA & 1.5M & 75.9 & 78.2 & 75.9 & 75.5\\
             DoRA & 1.6M & 76.9 & 78.7 & 76.9 & 76.2\\
             BiDoRA & 1.6M & \textbf{78.8} & \textbf{79.1} & \textbf{78.8} & \textbf{78.2}\\ \bottomrule
        \end{tabular}
        \label{tab:hot-protein}
    \end{minipage}\hfill
    \begin{minipage}{0.35\textwidth}
        \centering
        \begin{tabular}{c|cccc} \toprule
              \#Params & Accuracy & Precision & Recall & F1 \\ \midrule
              652.9M & 89.4 & 89.9 & 89.4 & 89.4\\ \midrule
              1.9M & 86.8 & 87.7 & 86.8 & 86.7\\
              2.0M &89.4 &91.3 &89.4 & 89.3\\
              2.0M & \textbf{92.1} & \textbf{93.1} & \textbf{92.1} & \textbf{92.0} \\ \bottomrule
        \end{tabular}
        \label{tab:bbp}
    \end{minipage}\hfill
    \begin{minipage}{0.18\textwidth}
        \centering
        \begin{tabular}{c|c} \toprule
             \#Params & MSE \\ \midrule
              652.7M & 0.2894 \\ \midrule
              1.7M & 0.3433 \\
              1.8M & 0.2918 \\
              1.8M & \textbf{0.2818} \\ \bottomrule
        \end{tabular}
        \label{tab:mic}
    \end{minipage}
\end{table}

We conduct additional experiments on biomedical datasets, including two classification tasks—thermostability prediction (\citet{chenhotprotein}, 936 training samples) and blood-brain barrier peptide prediction (BBP, \citet{dai2021bbppred}, 200 training samples)—and one regression task, minimum inhibitory concentration prediction (MIC, \citet{ledesma2023engineered}, 3,695 training samples), which contain significantly fewer samples than standard NLP tasks.

The results are presented in \cref{tab:hot-protein}.
Consistent with our previous findings, BiDoRA effectively fine-tunes pre-trained models on extremely small datasets.
Our method outperforms the baselines by a larger margin as the dataset size decreases, confirming our previous conclusion that our method effectively combats the overfitting issue on various network architectures and diverse tasks.

\subsection{Ablation Studies}

\begin{table*}[t]
    \centering
    \caption{Ablation studies. We evaluate the performance of BiDoRA without retraining (w/o retraining), without BLO ($\xi=0$), without orthogonal regularization (w/o cst.), and with retraining magnitude. 
    }
    \vspace{0.2cm}
    \setlength{\tabcolsep}{2pt}
    \begin{tabular}{c|cccccccccc} \toprule
        Method &  MNLI & SST-2 & MRPC & CoLA & QNLI & QQP & RTE & STS-B & Avg. \\ \midrule
        BiDoRA (retraining magnitude)&$87.0$&$94.3$&$89.1$&$60.7$&$\textbf{92.7}$& $91.0$ & $73.4$&$89.9$&$84.8$\\
        BiDoRA (w/o retraining) &$87.0$&$94.2$&$89.0$&$57.3$&$92.4$&$90.6$&$71.6$&$90.0$ & $84.0$\\
        BiDoRA ($\xi = 0$) &$86.9$ &$94.2$ & $89.0$& $59.4$&  $90.8$& $\textbf{91.2}$& $75.9$& $90.0$ &  84.7\\
        BiDoRA (w/o cst.)  & $87.0$ & $\textbf{94.4}$ & $88.6$ & $\textbf{61.3}$ & $\textbf{92.7}$ & $90.2$ & $76.0$ & $\textbf{90.1}$ & $ 85.0$ \\
        BiDoRA & $\textbf{87.1}$ & $\textbf{94.4}$ & $\textbf{89.4}$ & $\textbf{61.3}$ & $\textbf{92.7}$ & $90.6$ & $\textbf{76.1}$ & $\textbf{90.1}$ & $\textbf{85.2}$\\ \bottomrule
\end{tabular}
    \label{tab:ablation}
\end{table*}

In this section, we perform ablation studies to investigate the effectiveness of individual modules or strategies in BiDoRA. We fine-tune a RoBERTa-base model on the GLUE benchmark under different ablation settings, and the results are shown in \cref{tab:ablation}. 

\paragraph{Retraining.}
We test the model directly obtained from the search phase to evaluate the effectiveness of further retraining the direction component.
The results show that BiDoRA outperforms BiDoRA (w/o retraining) on average, highlighting the necessity of retraining.
\cref{tab:ablation} also validates that retraining the direction component leads to superior performance than retraining the magnitude.

\paragraph{Bi-level optimization.}
We set $\xi$ to zero in \cref{alg:bidora} to assess the effectiveness of the BLO framework.
This ablation setting can be interpreted as an alternative learning method where two optimization steps are carried out alternately on two different splits of the training dataset.
Notably, in the alternative learning method, the updating of each component is unaware of the others, making the training less stable.
In contrast, the hyper-gradient used in BLO avoids this issue by connecting the two levels in a certain way. 
The results show that BiDoRA outperforms BiDoRA ($\xi = 0$) on average, demonstrating the efficacy of the BLo strategy.

\paragraph{Orthogonal regularization.}
We examine the effectiveness of the orthogonality constraint in \cref{eq:regularizer} by setting $\gamma$ to zero.
Results show that BiDoRA outperforms BiDoRA (w/o cst.) on average, indicating the effectiveness of applying the orthogonality regularizer to alleviate overfitting.

\subsection{Weight Decomposition Analysis}
\label{sec:wda}

\begin{figure*}
    \centering
    \subfigure[Full FT($k = -65.816$)]{\includegraphics[width=0.245\textwidth]{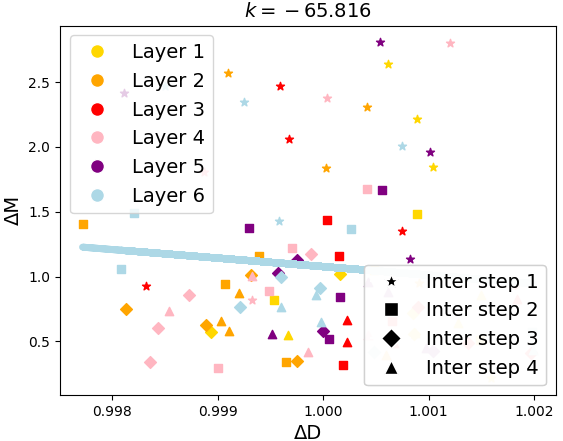}}
    \hspace{-0.01\textwidth} %
    \subfigure[LoRA($k = 0.836$)]{\includegraphics[width=0.25\textwidth]{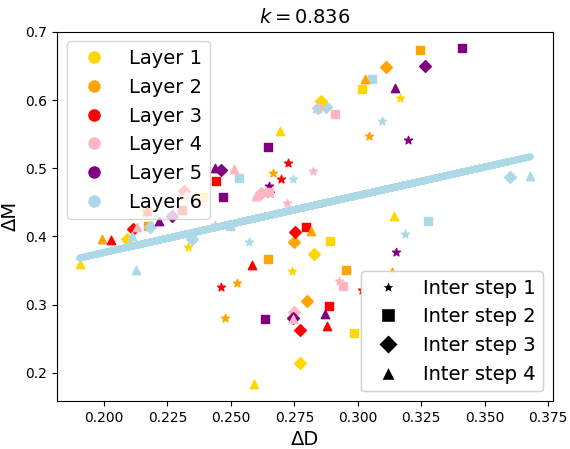}}
    \hspace{-0.01\textwidth}
    \subfigure[DoRA($k = -1.784$)]{\includegraphics[width=0.245\textwidth]{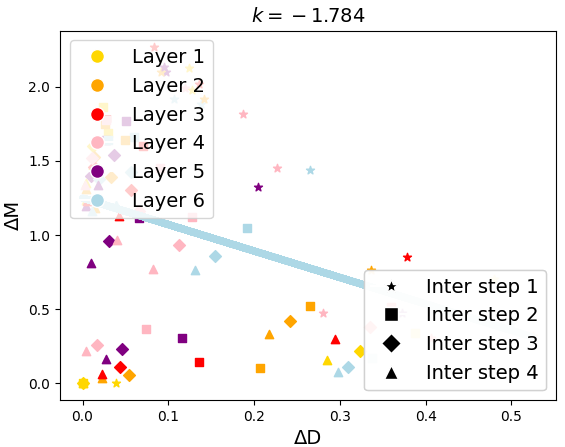}}
    \hspace{-0.01\textwidth}
    \subfigure[BiDoRA($k = -8.042$)]{\includegraphics[width=0.25\textwidth]{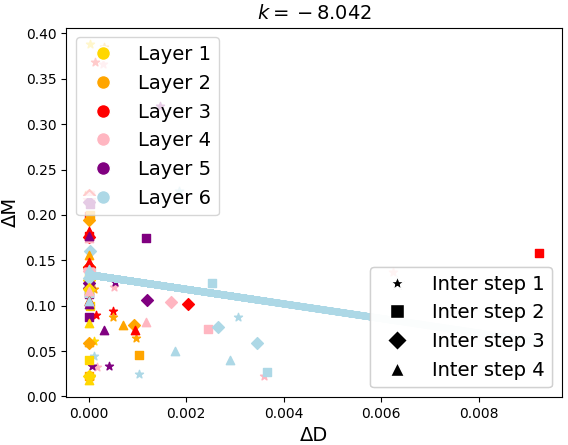}}
    \caption{Magnitude and direction updates for (a) FT, (b) LoRA, (c) DoRA, and (d) BiDoRA of the query matrices across different layers and intermediate steps after fine-tuning the GPT2 model on the E2E dataset \citep{novikova2017e2e}, where $k$ denotes the correlation value.
    Different markers represent matrices from different training steps, with each color corresponding to a specific layer.
    $\Delta \mathbf{M}$ denotes the average change in weight vector magnitude, and $\Delta \mathbf{D}$ denotes the average change in direction, as formally defined in \cref{sec:decomposition}.}

    \label{fig:wdaquery}
\end{figure*}

One important motivation of DoRA is to bridge the inherent differences between LoRA and FT.
Similar to DoRA, we conduct a weight decomposition analysis on the correlation between the change of magnitudes and that of directions (detailed in \cref{sec:decomposition}) for BiDoRA and baseline methods by fine-tuning a GPT2-medium model on the E2E dataset.
As shown in \cref{fig:wdaquery}, FT, DoRA, and BiDoRA all exhibit negative correlation values, while LoRA shows a positive correlation, consistent with the findings in \citet{liu2024dora}.
Notably, BiDoRA achieves a negative correlation of $-8.042$, closer to FT than DoRA's $-1.784$.
This improvement is attributed to the decoupled training process of the two layers, which allows for a higher learning capacity compared to DoRA.

\subsection{Discussion}
The advantage of BiDoRA is supported by both theoretical insights and empirical evidence, as detailed as follows.
\paragraph{Motivation.}
Theoretically, \citet{liu2024dora} showed that LoRA's training pattern tends to be coupled in terms of magnitude-direction correlation, which degrades learning capacity. Their solution was to introduce a reparameterization that decouples these components in the formulation.
We build upon DoRA following their theory and further decouple magnitude and direction in terms of training dynamics. Specifically, the two components are trained in separate loops within a bilevel optimization framework, which is expected to improve performance in an intuition similar to DoRA.

\paragraph{Empirical evidences.}    
We performed a Wilcoxon signed-rank test to compare the performance of DoRA and BiDoRA.
Specifically, we used the results from \cref{tab:nlu}.
For each PEFT method, we collected $9$ values ($8$ values from each dataset plus the average performance) from one base model. We concatenated the results from three base models (RoBERTa-base, RoBERTa-large, and DeBERTa-XXL) to obtain a list of $27$ values.
A comparison of these $27$ values between DoRA and BiDoRA reveals that BiDoRA is significantly better than DoRA, with a p-value of $2.4 \times 10^{-4}$.
This result demonstrates that BiDoRA offers a non-marginal improvement over DoRA.

Additionally, the weight decomposition analysis, including (\cref{fig:wdaquery} and \cref{fig:wdavalue}), indicates that BiDoRA achieves better decoupling of the components compared to DoRA.
Evaluation metrics across various tasks demonstrate the superior performance of BiDoRA, confirming that our decoupled optimization loop leads to improved outcomes.

\section{Conclusion and Future Works}
We propose BiDoRA, a novel bi-level optimization framework for PEFT of large-scale pre-trained models.
By conducting weight decomposition following the DoRA approach, our method trains the two components separately in two interconnected optimization levels using different sub-datasets.
In this way, BiDoRA not only decouples the learning process of the two components, resulting in a learning pattern closer to FT, but also effectively alleviates overfitting.
Empirical studies on various NLP tasks demonstrate that BiDoRA outperforms DoRA and other baselines, highlighting the effectiveness of our method.

One potential limitation of BiDoRA is its training efficiency (see \cref{sec:cost}) in terms of per-step cost, which could be reduced by using more advanced hyper-gradient estimators, such as SAMA \citep{choe2024making} or MixFlow-MG \citep{kemaev2025scalable}.
Furthermore, while we have empirically shown that BiDoRA induces better decoupling between the magnitude and direction components (\cref{fig:wdaquery} and \cref{fig:wdavalue}), a formal theoretical analysis of this property is currently lacking and serves for future work.

\subsubsection*{Acknowledgments}
P.X. acknowledges funding support from NSF IIS2405974, NSF IIS2339216, NIH R35GM157217, and NIH R21GM154171.

\bibliography{main}

\begin{thebibliography}{63}
\providecommand{\natexlab}[1]{#1}
\providecommand{\url}[1]{\texttt{#1}}
\expandafter\ifx\csname urlstyle\endcsname\relax
  \providecommand{\doi}[1]{doi: #1}\else
  \providecommand{\doi}{doi: \begingroup \urlstyle{rm}\Url}\fi

\bibitem[Azizi et~al.(2024)Azizi, Kundu, and Pedram]{azizi2024lamda}
Seyedarmin Azizi, Souvik Kundu, and Massoud Pedram.
\newblock Lamda: Large model fine-tuning via spectrally decomposed low-dimensional adaptation.
\newblock \emph{ArXiv preprint}, abs/2406.12832, 2024.
\newblock URL \url{https://arxiv.org/abs/2406.12832}.

\bibitem[Balestriero \& richard baraniuk(2018)Balestriero and richard baraniuk]{pmlr-v80-balestriero18b}
Randall Balestriero and richard baraniuk.
\newblock A spline theory of deep learning.
\newblock In Jennifer Dy and Andreas Krause (eds.), \emph{Proceedings of the 35th International Conference on Machine Learning}, volume~80 of \emph{Proceedings of Machine Learning Research}, pp.\  374--383. PMLR, 2018.
\newblock URL \url{https://proceedings.mlr.press/v80/balestriero18b.html}.

\bibitem[Banerjee \& Lavie(2005)Banerjee and Lavie]{banerjee2005meteor}
Satanjeev Banerjee and Alon Lavie.
\newblock {METEOR}: An automatic metric for {MT} evaluation with improved correlation with human judgments.
\newblock In Jade Goldstein, Alon Lavie, Chin-Yew Lin, and Clare Voss (eds.), \emph{Proceedings of the {ACL} Workshop on Intrinsic and Extrinsic Evaluation Measures for Machine Translation and/or Summarization}, pp.\  65--72, Ann Arbor, Michigan, 2005. Association for Computational Linguistics.
\newblock URL \url{https://aclanthology.org/W05-0909}.

\bibitem[Bansal et~al.(2018)Bansal, Chen, and Wang]{Bansal2018CanWG}
Nitin Bansal, Xiaohan Chen, and Zhangyang Wang.
\newblock Can we gain more from orthogonality regularizations in training deep cnns?
\newblock In \emph{Neural Information Processing Systems}, 2018.
\newblock URL \url{https://api.semanticscholar.org/CorpusID:55704502}.

\bibitem[Bao et~al.(2021)Bao, Wu, Li, Zhu, and Zhang]{bao2021stability}
Fan Bao, Guoqiang Wu, Chongxuan Li, Jun Zhu, and Bo~Zhang.
\newblock Stability and generalization of bilevel programming in hyperparameter optimization.
\newblock In Marc'Aurelio Ranzato, Alina Beygelzimer, Yann~N. Dauphin, Percy Liang, and Jennifer~Wortman Vaughan (eds.), \emph{Advances in Neural Information Processing Systems 34: Annual Conference on Neural Information Processing Systems 2021, NeurIPS 2021, December 6-14, 2021, virtual}, pp.\  4529--4541, 2021.
\newblock URL \url{https://proceedings.neurips.cc/paper/2021/hash/2406a0a94c80406914ff2f6c9fdd67d5-Abstract.html}.

\bibitem[Bi et~al.(2024)Bi, Yi, Zheng, Zhan, Huang, Ji, Li, and Zheng]{bi2024learning}
Qi~Bi, Jingjun Yi, Hao Zheng, Haolan Zhan, Yawen Huang, Wei Ji, Yuexiang Li, and Yefeng Zheng.
\newblock Learning frequency-adapted vision foundation model for domain generalized semantic segmentation.
\newblock In Amir Globersons, Lester Mackey, Danielle Belgrave, Angela Fan, Ulrich Paquet, Jakub~M. Tomczak, and Cheng Zhang (eds.), \emph{Advances in Neural Information Processing Systems 38: Annual Conference on Neural Information Processing Systems 2024, NeurIPS 2024, Vancouver, BC, Canada, December 10 - 15, 2024}, 2024.
\newblock URL \url{http://papers.nips.cc/paper\_files/paper/2024/hash/aaf50c91c3fc018f6a476032d02114d9-Abstract-Conference.html}.

\bibitem[Brown et~al.(2020)Brown, Mann, Ryder, Subbiah, Kaplan, Dhariwal, Neelakantan, Shyam, Sastry, Askell, Agarwal, Herbert{-}Voss, Krueger, Henighan, Child, Ramesh, Ziegler, Wu, Winter, Hesse, Chen, Sigler, Litwin, Gray, Chess, Clark, Berner, McCandlish, Radford, Sutskever, and Amodei]{brown2020language}
Tom~B. Brown, Benjamin Mann, Nick Ryder, Melanie Subbiah, Jared Kaplan, Prafulla Dhariwal, Arvind Neelakantan, Pranav Shyam, Girish Sastry, Amanda Askell, Sandhini Agarwal, Ariel Herbert{-}Voss, Gretchen Krueger, Tom Henighan, Rewon Child, Aditya Ramesh, Daniel~M. Ziegler, Jeffrey Wu, Clemens Winter, Christopher Hesse, Mark Chen, Eric Sigler, Mateusz Litwin, Scott Gray, Benjamin Chess, Jack Clark, Christopher Berner, Sam McCandlish, Alec Radford, Ilya Sutskever, and Dario Amodei.
\newblock Language models are few-shot learners.
\newblock In Hugo Larochelle, Marc'Aurelio Ranzato, Raia Hadsell, Maria{-}Florina Balcan, and Hsuan{-}Tien Lin (eds.), \emph{Advances in Neural Information Processing Systems 33: Annual Conference on Neural Information Processing Systems 2020, NeurIPS 2020, December 6-12, 2020, virtual}, 2020.
\newblock URL \url{https://proceedings.neurips.cc/paper/2020/hash/1457c0d6bfcb4967418bfb8ac142f64a-Abstract.html}.

\bibitem[Cer et~al.(2017)Cer, Diab, Agirre, Lopez-Gazpio, and Specia]{cer2017semeval}
Daniel Cer, Mona Diab, Eneko Agirre, I{\~n}igo Lopez-Gazpio, and Lucia Specia.
\newblock {S}em{E}val-2017 task 1: Semantic textual similarity multilingual and crosslingual focused evaluation.
\newblock In Steven Bethard, Marine Carpuat, Marianna Apidianaki, Saif~M. Mohammad, Daniel Cer, and David Jurgens (eds.), \emph{Proceedings of the 11th International Workshop on Semantic Evaluation ({S}em{E}val-2017)}, pp.\  1--14, Vancouver, Canada, 2017. Association for Computational Linguistics.
\newblock \doi{10.18653/v1/S17-2001}.
\newblock URL \url{https://aclanthology.org/S17-2001}.

\bibitem[Chen et~al.(2023)Chen, Gong, Diaz, Chen, Wells, Liu, Wang, Ellington, Dimakis, and Klivans]{chenhotprotein}
Tianlong Chen, Chengyue Gong, Daniel~Jesus Diaz, Xuxi Chen, Jordan~Tyler Wells, Qiang Liu, Zhangyang Wang, Andrew~D. Ellington, Alex Dimakis, and Adam~R. Klivans.
\newblock Hotprotein: {A} novel framework for protein thermostability prediction and editing.
\newblock In \emph{The Eleventh International Conference on Learning Representations, {ICLR} 2023, Kigali, Rwanda, May 1-5, 2023}. OpenReview.net, 2023.
\newblock URL \url{https://openreview.net/pdf?id=YDJRFWBMNby}.

\bibitem[Choe et~al.(2023{\natexlab{a}})Choe, Mehta, Ahn, Neiswanger, Xie, Strubell, and Xing]{choe2024making}
Sang~Keun Choe, Sanket~Vaibhav Mehta, Hwijeen Ahn, Willie Neiswanger, Pengtao Xie, Emma Strubell, and Eric~P. Xing.
\newblock Making scalable meta learning practical.
\newblock In Alice Oh, Tristan Naumann, Amir Globerson, Kate Saenko, Moritz Hardt, and Sergey Levine (eds.), \emph{Advances in Neural Information Processing Systems 36: Annual Conference on Neural Information Processing Systems 2023, NeurIPS 2023, New Orleans, LA, USA, December 10 - 16, 2023}, 2023{\natexlab{a}}.
\newblock URL \url{http://papers.nips.cc/paper\_files/paper/2023/hash/531998dc1fc858b5857a90b74d96ecab-Abstract-Conference.html}.

\bibitem[Choe et~al.(2023{\natexlab{b}})Choe, Neiswanger, Xie, and Xing]{choebetty}
Sang~Keun Choe, Willie Neiswanger, Pengtao Xie, and Eric~P. Xing.
\newblock Betty: An automatic differentiation library for multilevel optimization.
\newblock In \emph{The Eleventh International Conference on Learning Representations, {ICLR} 2023, Kigali, Rwanda, May 1-5, 2023}. OpenReview.net, 2023{\natexlab{b}}.
\newblock URL \url{https://openreview.net/pdf?id=LV\_MeMS38Q9}.

\bibitem[Collier et~al.(2004)Collier, Ohta, Tsuruoka, Tateisi, and Kim]{collier2004introduction}
Nigel Collier, Tomoko Ohta, Yoshimasa Tsuruoka, Yuka Tateisi, and Jin-Dong Kim.
\newblock Introduction to the bio-entity recognition task at {JNLPBA}.
\newblock In Nigel Collier, Patrick Ruch, and Adeline Nazarenko (eds.), \emph{Proceedings of the International Joint Workshop on Natural Language Processing in Biomedicine and its Applications ({NLPBA}/{B}io{NLP})}, pp.\  73--78, Geneva, Switzerland, 2004. COLING.
\newblock URL \url{https://aclanthology.org/W04-1213}.

\bibitem[Cui \& Bai(2019)Cui and Bai]{cui2019new}
Hua Cui and Jie Bai.
\newblock A new hyperparameters optimization method for convolutional neural networks.
\newblock \emph{Pattern Recognition Letters}, 125:\penalty0 828--834, 2019.

\bibitem[Dagan et~al.(2005)Dagan, Glickman, and Magnini]{dagan2005pascal}
Ido Dagan, Oren Glickman, and Bernardo Magnini.
\newblock The pascal recognising textual entailment challenge.
\newblock In \emph{Machine learning challenges workshop}, pp.\  177--190. Springer, 2005.

\bibitem[Dai et~al.(2021)Dai, Zhang, Tang, Wynendaele, Zhu, Bin, De~Spiegeleer, and Xia]{dai2021bbppred}
Ruyu Dai, Wei Zhang, Wending Tang, Evelien Wynendaele, Qizhi Zhu, Yannan Bin, Bart De~Spiegeleer, and Junfeng Xia.
\newblock Bbppred: sequence-based prediction of blood-brain barrier peptides with feature representation learning and logistic regression.
\newblock \emph{Journal of Chemical Information and Modeling}, 61\penalty0 (1):\penalty0 525--534, 2021.

\bibitem[Dolan \& Brockett(2005)Dolan and Brockett]{dolan2005automatically}
William~B. Dolan and Chris Brockett.
\newblock Automatically constructing a corpus of sentential paraphrases.
\newblock In \emph{Proceedings of the Third International Workshop on Paraphrasing ({IWP}2005)}, 2005.
\newblock URL \url{https://aclanthology.org/I05-5002}.

\bibitem[Finn et~al.(2017)Finn, Abbeel, and Levine]{finn2017model}
Chelsea Finn, Pieter Abbeel, and Sergey Levine.
\newblock Model-agnostic meta-learning for fast adaptation of deep networks.
\newblock In Doina Precup and Yee~Whye Teh (eds.), \emph{Proceedings of the 34th International Conference on Machine Learning, {ICML} 2017, Sydney, NSW, Australia, 6-11 August 2017}, volume~70 of \emph{Proceedings of Machine Learning Research}, pp.\  1126--1135. {PMLR}, 2017.
\newblock URL \url{http://proceedings.mlr.press/v70/finn17a.html}.

\bibitem[Franceschi et~al.(2017)Franceschi, Donini, Frasconi, and Pontil]{franceschi2017forward}
Luca Franceschi, Michele Donini, Paolo Frasconi, and Massimiliano Pontil.
\newblock Forward and reverse gradient-based hyperparameter optimization.
\newblock In Doina Precup and Yee~Whye Teh (eds.), \emph{Proceedings of the 34th International Conference on Machine Learning, {ICML} 2017, Sydney, NSW, Australia, 6-11 August 2017}, volume~70 of \emph{Proceedings of Machine Learning Research}, pp.\  1165--1173. {PMLR}, 2017.
\newblock URL \url{http://proceedings.mlr.press/v70/franceschi17a.html}.

\bibitem[Gao et~al.(2024)Gao, Wang, Chen, Liu, Wu, Chen, and Li]{gao2024parameter}
Ziqi Gao, Qichao Wang, Aochuan Chen, Zijing Liu, Bingzhe Wu, Liang Chen, and Jia Li.
\newblock Parameter-efficient fine-tuning with discrete fourier transform.
\newblock In \emph{Forty-first International Conference on Machine Learning, {ICML} 2024, Vienna, Austria, July 21-27, 2024}. OpenReview.net, 2024.
\newblock URL \url{https://openreview.net/forum?id=XUOHKSsurt}.

\bibitem[He et~al.(2022)He, Zhou, Ma, Berg{-}Kirkpatrick, and Neubig]{he2021towards}
Junxian He, Chunting Zhou, Xuezhe Ma, Taylor Berg{-}Kirkpatrick, and Graham Neubig.
\newblock Towards a unified view of parameter-efficient transfer learning.
\newblock In \emph{The Tenth International Conference on Learning Representations, {ICLR} 2022, Virtual Event, April 25-29, 2022}. OpenReview.net, 2022.
\newblock URL \url{https://openreview.net/forum?id=0RDcd5Axok}.

\bibitem[Houlsby et~al.(2019)Houlsby, Giurgiu, Jastrzebski, Morrone, de~Laroussilhe, Gesmundo, Attariyan, and Gelly]{houlsby2019parameter}
Neil Houlsby, Andrei Giurgiu, Stanislaw Jastrzebski, Bruna Morrone, Quentin de~Laroussilhe, Andrea Gesmundo, Mona Attariyan, and Sylvain Gelly.
\newblock Parameter-efficient transfer learning for {NLP}.
\newblock In Kamalika Chaudhuri and Ruslan Salakhutdinov (eds.), \emph{Proceedings of the 36th International Conference on Machine Learning, {ICML} 2019, 9-15 June 2019, Long Beach, California, {USA}}, volume~97 of \emph{Proceedings of Machine Learning Research}, pp.\  2790--2799. {PMLR}, 2019.
\newblock URL \url{http://proceedings.mlr.press/v97/houlsby19a.html}.

\bibitem[Hu et~al.(2022{\natexlab{a}})Hu, Shen, Wallis, Allen{-}Zhu, Li, Wang, Wang, and Chen]{hu2022lora}
Edward~J. Hu, Yelong Shen, Phillip Wallis, Zeyuan Allen{-}Zhu, Yuanzhi Li, Shean Wang, Lu~Wang, and Weizhu Chen.
\newblock Lora: Low-rank adaptation of large language models.
\newblock In \emph{The Tenth International Conference on Learning Representations, {ICLR} 2022, Virtual Event, April 25-29, 2022}. OpenReview.net, 2022{\natexlab{a}}.
\newblock URL \url{https://openreview.net/forum?id=nZeVKeeFYf9}.

\bibitem[Hu et~al.(2022{\natexlab{b}})Hu, Shen, Wallis, Allen{-}Zhu, Li, Wang, Wang, and Chen]{hulora}
Edward~J. Hu, Yelong Shen, Phillip Wallis, Zeyuan Allen{-}Zhu, Yuanzhi Li, Shean Wang, Lu~Wang, and Weizhu Chen.
\newblock Lora: Low-rank adaptation of large language models.
\newblock In \emph{The Tenth International Conference on Learning Representations, {ICLR} 2022, Virtual Event, April 25-29, 2022}. OpenReview.net, 2022{\natexlab{b}}.
\newblock URL \url{https://openreview.net/forum?id=nZeVKeeFYf9}.

\bibitem[Kemaev et~al.(2025)Kemaev, Calian, Zintgraf, Farquhar, and van Hasselt]{kemaev2025scalable}
Iurii Kemaev, Dan~A Calian, Luisa~M Zintgraf, Gregory Farquhar, and Hado van Hasselt.
\newblock Scalable meta-learning via mixed-mode differentiation.
\newblock \emph{ArXiv preprint}, abs/2505.00793, 2025.
\newblock URL \url{https://arxiv.org/abs/2505.00793}.

\bibitem[Kopiczko et~al.(2024)Kopiczko, Blankevoort, and Asano]{kopiczko2023vera}
Dawid~Jan Kopiczko, Tijmen Blankevoort, and Yuki~M. Asano.
\newblock Vera: Vector-based random matrix adaptation.
\newblock In \emph{The Twelfth International Conference on Learning Representations, {ICLR} 2024, Vienna, Austria, May 7-11, 2024}. OpenReview.net, 2024.
\newblock URL \url{https://openreview.net/forum?id=NjNfLdxr3A}.

\bibitem[Ledesma-Fernandez et~al.(2023)Ledesma-Fernandez, Velasco-Lozano, Santiago-Arcos, L{\'o}pez-Gallego, and Cortajarena]{ledesma2023engineered}
Alba Ledesma-Fernandez, Susana Velasco-Lozano, Javier Santiago-Arcos, Fernando L{\'o}pez-Gallego, and Aitziber~L Cortajarena.
\newblock Engineered repeat proteins as scaffolds to assemble multi-enzyme systems for efficient cell-free biosynthesis.
\newblock \emph{Nature Communications}, 14\penalty0 (1):\penalty0 2587, 2023.

\bibitem[Lester et~al.(2021)Lester, Al-Rfou, and Constant]{lester2021power}
Brian Lester, Rami Al-Rfou, and Noah Constant.
\newblock The power of scale for parameter-efficient prompt tuning.
\newblock In Marie-Francine Moens, Xuanjing Huang, Lucia Specia, and Scott Wen-tau Yih (eds.), \emph{Proceedings of the 2021 Conference on Empirical Methods in Natural Language Processing}, pp.\  3045--3059, Online and Punta Cana, Dominican Republic, 2021. Association for Computational Linguistics.
\newblock \doi{10.18653/v1/2021.emnlp-main.243}.
\newblock URL \url{https://aclanthology.org/2021.emnlp-main.243}.

\bibitem[Lin(2004)]{lin2004rouge}
Chin-Yew Lin.
\newblock {ROUGE}: A package for automatic evaluation of summaries.
\newblock In \emph{Text Summarization Branches Out}, pp.\  74--81, Barcelona, Spain, 2004. Association for Computational Linguistics.
\newblock URL \url{https://aclanthology.org/W04-1013}.

\bibitem[Lin \& Och(2004)Lin and Och]{lin2004automatic}
Chin-Yew Lin and Franz~Josef Och.
\newblock Automatic evaluation of machine translation quality using longest common subsequence and skip-bigram statistics.
\newblock In \emph{Proceedings of the 42nd Annual Meeting of the Association for Computational Linguistics ({ACL}-04)}, pp.\  605--612, Barcelona, Spain, 2004.
\newblock \doi{10.3115/1218955.1219032}.
\newblock URL \url{https://aclanthology.org/P04-1077}.

\bibitem[Lin et~al.(2020)Lin, Madotto, and Fung]{lin2020exploring}
Zhaojiang Lin, Andrea Madotto, and Pascale Fung.
\newblock Exploring versatile generative language model via parameter-efficient transfer learning.
\newblock In Trevor Cohn, Yulan He, and Yang Liu (eds.), \emph{Findings of the Association for Computational Linguistics: EMNLP 2020}, pp.\  441--459, Online, 2020. Association for Computational Linguistics.
\newblock \doi{10.18653/v1/2020.findings-emnlp.41}.
\newblock URL \url{https://aclanthology.org/2020.findings-emnlp.41}.

\bibitem[Liu et~al.(2019)Liu, Simonyan, and Yang]{liu2018darts}
Hanxiao Liu, Karen Simonyan, and Yiming Yang.
\newblock {DARTS:} differentiable architecture search.
\newblock In \emph{7th International Conference on Learning Representations, {ICLR} 2019, New Orleans, LA, USA, May 6-9, 2019}. OpenReview.net, 2019.
\newblock URL \url{https://openreview.net/forum?id=S1eYHoC5FX}.

\bibitem[Liu et~al.(2024{\natexlab{a}})Liu, Wang, Yin, Molchanov, Wang, Cheng, and Chen]{liu2024dora}
Shih{-}Yang Liu, Chien{-}Yi Wang, Hongxu Yin, Pavlo Molchanov, Yu{-}Chiang~Frank Wang, Kwang{-}Ting Cheng, and Min{-}Hung Chen.
\newblock Dora: Weight-decomposed low-rank adaptation.
\newblock In \emph{Forty-first International Conference on Machine Learning, {ICML} 2024, Vienna, Austria, July 21-27, 2024}. OpenReview.net, 2024{\natexlab{a}}.
\newblock URL \url{https://openreview.net/forum?id=3d5CIRG1n2}.

\bibitem[Liu et~al.(2024{\natexlab{b}})Liu, Kundu, Li, Wan, Jiang, and Beerel]{liu2024aflora}
Zeyu Liu, Souvik Kundu, Anni Li, Junrui Wan, Lianghao Jiang, and Peter~Anthony Beerel.
\newblock Aflora: Adaptive freezing of low rank adaptation in parameter efficient fine-tuning of large models.
\newblock \emph{ArXiv preprint}, abs/2403.13269, 2024{\natexlab{b}}.
\newblock URL \url{https://arxiv.org/abs/2403.13269}.

\bibitem[Lorraine et~al.(2020)Lorraine, Vicol, and Duvenaud]{lorraine2020optimizing}
Jonathan Lorraine, Paul Vicol, and David Duvenaud.
\newblock Optimizing millions of hyperparameters by implicit differentiation.
\newblock In Silvia Chiappa and Roberto Calandra (eds.), \emph{The 23rd International Conference on Artificial Intelligence and Statistics, {AISTATS} 2020, 26-28 August 2020, Online [Palermo, Sicily, Italy]}, volume 108 of \emph{Proceedings of Machine Learning Research}, pp.\  1540--1552. {PMLR}, 2020.
\newblock URL \url{http://proceedings.mlr.press/v108/lorraine20a.html}.

\bibitem[Novikova et~al.(2017)Novikova, Du{\v{s}}ek, and Rieser]{novikova2017e2e}
Jekaterina Novikova, Ond{\v{r}}ej Du{\v{s}}ek, and Verena Rieser.
\newblock The {E}2{E} dataset: New challenges for end-to-end generation.
\newblock In Kristiina Jokinen, Manfred Stede, David DeVault, and Annie Louis (eds.), \emph{Proceedings of the 18th Annual {SIG}dial Meeting on Discourse and Dialogue}, pp.\  201--206, Saarbr{\"u}cken, Germany, 2017. Association for Computational Linguistics.
\newblock \doi{10.18653/v1/W17-5525}.
\newblock URL \url{https://aclanthology.org/W17-5525}.

\bibitem[Padmanabhan et~al.(2016)Padmanabhan, Bhat, Shevade, and Narahari]{padmanabhan2016topic}
Divya Padmanabhan, Satyanath Bhat, Shirish Shevade, and Y~Narahari.
\newblock Topic model based multi-label classification.
\newblock In \emph{2016 IEEE 28th International Conference on Tools with Artificial Intelligence (ICTAI)}, pp.\  996--1003. IEEE, 2016.

\bibitem[Papineni et~al.(2002)Papineni, Roukos, Ward, and Zhu]{papineni2002bleu}
Kishore Papineni, Salim Roukos, Todd Ward, and Wei-Jing Zhu.
\newblock {B}leu: a method for automatic evaluation of machine translation.
\newblock In Pierre Isabelle, Eugene Charniak, and Dekang Lin (eds.), \emph{Proceedings of the 40th Annual Meeting of the Association for Computational Linguistics}, pp.\  311--318, Philadelphia, Pennsylvania, USA, 2002. Association for Computational Linguistics.
\newblock \doi{10.3115/1073083.1073135}.
\newblock URL \url{https://aclanthology.org/P02-1040}.

\bibitem[Pearlmutter \& Siskind(2008)Pearlmutter and Siskind]{pearlmutter2008reverse}
Barak~A Pearlmutter and Jeffrey~Mark Siskind.
\newblock Reverse-mode ad in a functional framework: Lambda the ultimate backpropagator.
\newblock \emph{ACM Transactions on Programming Languages and Systems (TOPLAS)}, 30\penalty0 (2):\penalty0 1--36, 2008.

\bibitem[Pedregosa(2016)]{pedregosa2016hyperparameter}
Fabian Pedregosa.
\newblock Hyperparameter optimization with approximate gradient.
\newblock In Maria{-}Florina Balcan and Kilian~Q. Weinberger (eds.), \emph{Proceedings of the 33nd International Conference on Machine Learning, {ICML} 2016, New York City, NY, USA, June 19-24, 2016}, volume~48 of \emph{{JMLR} Workshop and Conference Proceedings}, pp.\  737--746. JMLR.org, 2016.
\newblock URL \url{http://proceedings.mlr.press/v48/pedregosa16.html}.

\bibitem[Pfeiffer et~al.(2021)Pfeiffer, Kamath, R{\"u}ckl{\'e}, Cho, and Gurevych]{pfeiffer2020adapterfusion}
Jonas Pfeiffer, Aishwarya Kamath, Andreas R{\"u}ckl{\'e}, Kyunghyun Cho, and Iryna Gurevych.
\newblock {A}dapter{F}usion: Non-destructive task composition for transfer learning.
\newblock In Paola Merlo, Jorg Tiedemann, and Reut Tsarfaty (eds.), \emph{Proceedings of the 16th Conference of the European Chapter of the Association for Computational Linguistics: Main Volume}, pp.\  487--503, Online, 2021. Association for Computational Linguistics.
\newblock \doi{10.18653/v1/2021.eacl-main.39}.
\newblock URL \url{https://aclanthology.org/2021.eacl-main.39}.

\bibitem[Radford et~al.(2019)Radford, Wu, Child, Luan, Amodei, Sutskever, et~al.]{radford2019language}
Alec Radford, Jeffrey Wu, Rewon Child, David Luan, Dario Amodei, Ilya Sutskever, et~al.
\newblock Language models are unsupervised multitask learners.
\newblock \emph{OpenAI blog}, 1\penalty0 (8):\penalty0 9, 2019.

\bibitem[Rajeswaran et~al.(2019)Rajeswaran, Finn, Kakade, and Levine]{rajeswaran2019meta}
Aravind Rajeswaran, Chelsea Finn, Sham~M. Kakade, and Sergey Levine.
\newblock Meta-learning with implicit gradients.
\newblock In Hanna~M. Wallach, Hugo Larochelle, Alina Beygelzimer, Florence d'Alch{\'{e}}{-}Buc, Emily~B. Fox, and Roman Garnett (eds.), \emph{Advances in Neural Information Processing Systems 32: Annual Conference on Neural Information Processing Systems 2019, NeurIPS 2019, December 8-14, 2019, Vancouver, BC, Canada}, pp.\  113--124, 2019.
\newblock URL \url{https://proceedings.neurips.cc/paper/2019/hash/072b030ba126b2f4b2374f342be9ed44-Abstract.html}.

\bibitem[Rajpurkar et~al.(2018)Rajpurkar, Jia, and Liang]{rajpurkar2018know}
Pranav Rajpurkar, Robin Jia, and Percy Liang.
\newblock Know what you don{'}t know: Unanswerable questions for {SQ}u{AD}.
\newblock In Iryna Gurevych and Yusuke Miyao (eds.), \emph{Proceedings of the 56th Annual Meeting of the Association for Computational Linguistics (Volume 2: Short Papers)}, pp.\  784--789, Melbourne, Australia, 2018. Association for Computational Linguistics.
\newblock \doi{10.18653/v1/P18-2124}.
\newblock URL \url{https://aclanthology.org/P18-2124}.

\bibitem[Razdaibiedina et~al.(2023)Razdaibiedina, Mao, Khabsa, Lewis, Hou, Ba, and Almahairi]{razdaibiedina2023residual}
Anastasiia Razdaibiedina, Yuning Mao, Madian Khabsa, Mike Lewis, Rui Hou, Jimmy Ba, and Amjad Almahairi.
\newblock Residual prompt tuning: improving prompt tuning with residual reparameterization.
\newblock In Anna Rogers, Jordan Boyd-Graber, and Naoaki Okazaki (eds.), \emph{Findings of the Association for Computational Linguistics: ACL 2023}, pp.\  6740--6757, Toronto, Canada, 2023. Association for Computational Linguistics.
\newblock \doi{10.18653/v1/2023.findings-acl.421}.
\newblock URL \url{https://aclanthology.org/2023.findings-acl.421}.

\bibitem[Rives et~al.(2021)Rives, Meier, Sercu, Goyal, Lin, Liu, Guo, Ott, Zitnick, Ma, et~al.]{rives2021biological}
Alexander Rives, Joshua Meier, Tom Sercu, Siddharth Goyal, Zeming Lin, Jason Liu, Demi Guo, Myle Ott, C~Lawrence Zitnick, Jerry Ma, et~al.
\newblock Biological structure and function emerge from scaling unsupervised learning to 250 million protein sequences.
\newblock \emph{Proceedings of the National Academy of Sciences}, 118\penalty0 (15):\penalty0 e2016239118, 2021.

\bibitem[R{\"u}ckl{\'e} et~al.(2021)R{\"u}ckl{\'e}, Geigle, Glockner, Beck, Pfeiffer, Reimers, and Gurevych]{ruckle2020adapterdrop}
Andreas R{\"u}ckl{\'e}, Gregor Geigle, Max Glockner, Tilman Beck, Jonas Pfeiffer, Nils Reimers, and Iryna Gurevych.
\newblock {AdapterDrop}: {O}n the efficiency of adapters in transformers.
\newblock In Marie-Francine Moens, Xuanjing Huang, Lucia Specia, and Scott Wen-tau Yih (eds.), \emph{Proceedings of the 2021 Conference on Empirical Methods in Natural Language Processing}, pp.\  7930--7946, Online and Punta Cana, Dominican Republic, 2021. Association for Computational Linguistics.
\newblock \doi{10.18653/v1/2021.emnlp-main.626}.
\newblock URL \url{https://aclanthology.org/2021.emnlp-main.626}.

\bibitem[Shen et~al.(2025{\natexlab{a}})Shen, Bi, Huang, Zhu, Pimentel, and Pathania]{shen2025macp}
Yixian Shen, Qi~Bi, Jia-Hong Huang, Hongyi Zhu, Andy~D Pimentel, and Anuj Pathania.
\newblock Macp: Minimal yet mighty adaptation via hierarchical cosine projection.
\newblock \emph{ArXiv preprint}, abs/2505.23870, 2025{\natexlab{a}}.
\newblock URL \url{https://arxiv.org/abs/2505.23870}.

\bibitem[Shen et~al.(2025{\natexlab{b}})Shen, Bi, Huang, Zhu, Pimentel, and Pathania]{shen2025ssh}
Yixian Shen, Qi~Bi, Jia-hong Huang, Hongyi Zhu, Andy~D. Pimentel, and Anuj Pathania.
\newblock {SSH}: Sparse spectrum adaptation via discrete hartley transformation.
\newblock In Luis Chiruzzo, Alan Ritter, and Lu~Wang (eds.), \emph{Proceedings of the 2025 Conference of the Nations of the Americas Chapter of the Association for Computational Linguistics: Human Language Technologies (Volume 1: Long Papers)}, pp.\  10400--10415, Albuquerque, New Mexico, 2025{\natexlab{b}}. Association for Computational Linguistics.
\newblock ISBN 979-8-89176-189-6.
\newblock \doi{10.18653/v1/2025.naacl-long.522}.
\newblock URL \url{https://aclanthology.org/2025.naacl-long.522/}.

\bibitem[Socher et~al.(2013)Socher, Perelygin, Wu, Chuang, Manning, Ng, and Potts]{socher2013recursive}
Richard Socher, Alex Perelygin, Jean Wu, Jason Chuang, Christopher~D. Manning, Andrew Ng, and Christopher Potts.
\newblock Recursive deep models for semantic compositionality over a sentiment treebank.
\newblock In David Yarowsky, Timothy Baldwin, Anna Korhonen, Karen Livescu, and Steven Bethard (eds.), \emph{Proceedings of the 2013 Conference on Empirical Methods in Natural Language Processing}, pp.\  1631--1642, Seattle, Washington, USA, 2013. Association for Computational Linguistics.
\newblock URL \url{https://aclanthology.org/D13-1170}.

\bibitem[Tjong Kim~Sang(2002)]{sang2003introduction}
Erik~F. Tjong Kim~Sang.
\newblock Introduction to the {C}o{NLL}-2002 shared task: Language-independent named entity recognition.
\newblock In \emph{{COLING}-02: The 6th Conference on Natural Language Learning 2002 ({C}o{NLL}-2002)}, 2002.
\newblock URL \url{https://aclanthology.org/W02-2024}.

\bibitem[Vedantam et~al.(2015)Vedantam, Zitnick, and Parikh]{vedantam2015cider}
Ramakrishna Vedantam, C.~Lawrence Zitnick, and Devi Parikh.
\newblock Cider: Consensus-based image description evaluation.
\newblock In \emph{{IEEE} Conference on Computer Vision and Pattern Recognition, {CVPR} 2015, Boston, MA, USA, June 7-12, 2015}, pp.\  4566--4575. {IEEE} Computer Society, 2015.
\newblock \doi{10.1109/CVPR.2015.7299087}.
\newblock URL \url{https://doi.org/10.1109/CVPR.2015.7299087}.

\bibitem[Wang et~al.(2019)Wang, Singh, Michael, Hill, Levy, and Bowman]{wang2019glue}
Alex Wang, Amanpreet Singh, Julian Michael, Felix Hill, Omer Levy, and Samuel~R. Bowman.
\newblock {GLUE:} {A} multi-task benchmark and analysis platform for natural language understanding.
\newblock In \emph{7th International Conference on Learning Representations, {ICLR} 2019, New Orleans, LA, USA, May 6-9, 2019}. OpenReview.net, 2019.
\newblock URL \url{https://openreview.net/forum?id=rJ4km2R5t7}.

\bibitem[Wang et~al.(2017)Wang, Hamza, and Florian]{wang2017bilateral}
Zhiguo Wang, Wael Hamza, and Radu Florian.
\newblock Bilateral multi-perspective matching for natural language sentences.
\newblock In Carles Sierra (ed.), \emph{Proceedings of the Twenty-Sixth International Joint Conference on Artificial Intelligence, {IJCAI} 2017, Melbourne, Australia, August 19-25, 2017}, pp.\  4144--4150. ijcai.org, 2017.
\newblock \doi{10.24963/ijcai.2017/579}.
\newblock URL \url{https://doi.org/10.24963/ijcai.2017/579}.

\bibitem[Warstadt et~al.(2019)Warstadt, Singh, and Bowman]{warstadt2019neural}
Alex Warstadt, Amanpreet Singh, and Samuel~R. Bowman.
\newblock Neural network acceptability judgments.
\newblock \emph{Transactions of the Association for Computational Linguistics}, 7:\penalty0 625--641, 2019.
\newblock \doi{10.1162/tacl_a_00290}.
\newblock URL \url{https://aclanthology.org/Q19-1040}.

\bibitem[Williams et~al.(2018)Williams, Nangia, and Bowman]{williams2017broad}
Adina Williams, Nikita Nangia, and Samuel Bowman.
\newblock A broad-coverage challenge corpus for sentence understanding through inference.
\newblock In Marilyn Walker, Heng Ji, and Amanda Stent (eds.), \emph{Proceedings of the 2018 Conference of the North {A}merican Chapter of the Association for Computational Linguistics: Human Language Technologies, Volume 1 (Long Papers)}, pp.\  1112--1122, New Orleans, Louisiana, 2018. Association for Computational Linguistics.
\newblock \doi{10.18653/v1/N18-1101}.
\newblock URL \url{https://aclanthology.org/N18-1101}.

\bibitem[Wolf et~al.(2019)Wolf, Debut, Sanh, Chaumond, Delangue, Moi, Cistac, Rault, Louf, Funtowicz, et~al.]{wolf2019huggingface}
Thomas Wolf, Lysandre Debut, Victor Sanh, Julien Chaumond, Clement Delangue, Anthony Moi, Pierric Cistac, Tim Rault, R{\'e}mi Louf, Morgan Funtowicz, et~al.
\newblock Huggingface's transformers: State-of-the-art natural language processing.
\newblock \emph{ArXiv preprint}, abs/1910.03771, 2019.
\newblock URL \url{https://arxiv.org/abs/1910.03771}.

\bibitem[Xie et~al.(2017)Xie, Xiong, and Pu]{Xie2017gram}
Di~Xie, Jiang Xiong, and Shiliang Pu.
\newblock All you need is beyond a good init: Exploring better solution for training extremely deep convolutional neural networks with orthonormality and modulation.
\newblock In \emph{2017 {IEEE} Conference on Computer Vision and Pattern Recognition, {CVPR} 2017, Honolulu, HI, USA, July 21-26, 2017}, pp.\  5075--5084. {IEEE} Computer Society, 2017.
\newblock \doi{10.1109/CVPR.2017.539}.
\newblock URL \url{https://doi.org/10.1109/CVPR.2017.539}.

\bibitem[Xu et~al.(2023)Xu, Zhang, Wei, Hu, and Bai]{xu2023side}
Mengde Xu, Zheng Zhang, Fangyun Wei, Han Hu, and Xiang Bai.
\newblock Side adapter network for open-vocabulary semantic segmentation.
\newblock In \emph{{IEEE/CVF} Conference on Computer Vision and Pattern Recognition, {CVPR} 2023, Vancouver, BC, Canada, June 17-24, 2023}, pp.\  2945--2954. {IEEE}, 2023.
\newblock \doi{10.1109/CVPR52729.2023.00288}.
\newblock URL \url{https://doi.org/10.1109/CVPR52729.2023.00288}.

\bibitem[Yi et~al.(2024)Yi, Bi, Zheng, Zhan, Ji, Huang, Li, and Zheng]{yi2024learning}
Jingjun Yi, Qi~Bi, Hao Zheng, Haolan Zhan, Wei Ji, Yawen Huang, Yuexiang Li, and Yefeng Zheng.
\newblock Learning spectral-decomposited tokens for domain generalized semantic segmentation.
\newblock In \emph{Proceedings of the 32nd ACM International Conference on Multimedia}, pp.\  8159--8168, 2024.

\bibitem[Zhang et~al.(2024{\natexlab{a}})Zhang, Guo, Schaffer, Ko, Singh, Rahmani, Grotjahn, Villa, Gilson, Wang, et~al.]{zhang2024proteinaligner}
Li~Zhang, Han Guo, Leah~V Schaffer, Young~Su Ko, Digvijay Singh, Hamid Rahmani, Danielle Grotjahn, Elizabeth Villa, Michael Gilson, Wei Wang, et~al.
\newblock Proteinaligner: A multi-modal pretraining framework for protein foundation models.
\newblock \emph{bioRxiv}, pp.\  2024--10, 2024{\natexlab{a}}.

\bibitem[Zhang et~al.(2021)Zhang, Su, Pan, Chang, Abbasnejad, and Haffari]{zhang2021idarts}
Miao Zhang, Steven~W. Su, Shirui Pan, Xiaojun Chang, M.~Ehsan Abbasnejad, and Reza Haffari.
\newblock idarts: Differentiable architecture search with stochastic implicit gradients.
\newblock In Marina Meila and Tong Zhang (eds.), \emph{Proceedings of the 38th International Conference on Machine Learning, {ICML} 2021, 18-24 July 2021, Virtual Event}, volume 139 of \emph{Proceedings of Machine Learning Research}, pp.\  12557--12566. {PMLR}, 2021.
\newblock URL \url{http://proceedings.mlr.press/v139/zhang21s.html}.

\bibitem[Zhang et~al.(2023)Zhang, Chen, Bukharin, He, Cheng, Chen, and Zhao]{zhang2023adaptive}
Qingru Zhang, Minshuo Chen, Alexander Bukharin, Pengcheng He, Yu~Cheng, Weizhu Chen, and Tuo Zhao.
\newblock Adaptive budget allocation for parameter-efficient fine-tuning.
\newblock In \emph{The Eleventh International Conference on Learning Representations, {ICLR} 2023, Kigali, Rwanda, May 1-5, 2023}. OpenReview.net, 2023.
\newblock URL \url{https://openreview.net/pdf?id=lq62uWRJjiY}.

\bibitem[Zhang et~al.(2024{\natexlab{b}})Zhang, Qiang, Somayajula, and Xie]{zhang-etal-2024-autolora}
Ruiyi Zhang, Rushi Qiang, Sai~Ashish Somayajula, and Pengtao Xie.
\newblock {A}uto{L}o{RA}: Automatically tuning matrix ranks in low-rank adaptation based on meta learning.
\newblock In Kevin Duh, Helena Gomez, and Steven Bethard (eds.), \emph{Proceedings of the 2024 Conference of the North American Chapter of the Association for Computational Linguistics: Human Language Technologies (Volume 1: Long Papers)}, pp.\  5048--5060, Mexico City, Mexico, 2024{\natexlab{b}}. Association for Computational Linguistics.
\newblock URL \url{https://aclanthology.org/2024.naacl-long.282}.

\end{thebibliography}
\bibliographystyle{tmlr}

\newpage
\appendix

\section{Datasets and Models}
\label{sec:appendixdatasetsmodels}

\begin{table}[htbp]
\centering
\caption{Summary of datasets used in the experiments}
\label{tab:dataset_summary}
\vspace{0.2cm}
\begin{tabular}{@{}cc>{\raggedright\arraybackslash}p{3cm}ccc@{}}
\toprule
\textbf{Task Group} & \textbf{Dataset} & \textbf{Metrics} & \textbf{Train} & \textbf{Dev / Val} & \textbf{Test} \\
\midrule
\multirow{8}{*}{\makecell[c]{Natural Language\\Understanding}}
& MNLI & Accuracy & 393k & 20k & 20k \\
& SST-2 & Accuracy & 67k & 872 & 1.8k  \\
& MRPC & Accuracy & 3.7k & 408 & 1.7k  \\
& CoLA & Matthews Corr & 8.5k & 1k & 1k \\
& QNLI & Accuracy & 108k & 5.7k & 5.7k  \\
& QQP & Accuracy & 364k & 40k & 391k \\
& RTE & Accuracy & 2.5k & 276 & 3k \\
& STS-B & Pearson Corr & 7.0k & 1.5k & 1.4k  \\
\midrule
\multirow{3}{*}{\makecell[c]{Text\\Classification}}
& ModApte & F1 & 8.8k & - & 3k  \\
& ModHayes & F1 & 18k & - & 0.7k  \\
& ModLewis & F1 & 12k & - & 5.5k  \\
\midrule
\makecell[l]{Natural Language\\Generation} & E2E & \makecell[l]{BLEU, NIST, MET,\\ROUGE-L, CIDEr} & 42k & 4.6k & - \\
\midrule
\multirow{2}{*}{\makecell[c]{Token\\Classification}}
& BioNLP & \makecell[l]{Accuracy, Precision,\\Recall, F1} & 17k & 1.9k & 3.9k \\
& CoNLL2003 & \makecell[l]{Accuracy, Precision,\\Recall, F1} & 14k & 3.3k & 3.5k  \\

\midrule
\multirow{3}{*}{\makecell[c]{Biomedical \\ Experiments}}
& \makecell[c]{Thermostability \\ prediction} & \makecell[l]{Accuracy, Precision,\\Recall, F1} & 3,695 & - & 924  \\
& BBP & \makecell[l]{Accuracy, Precision,\\Recall, F1} & 200 & - & 38  \\
& MIC & \makecell[l]{MSE} & 936 & - & 104  \\

\bottomrule
\end{tabular}
\end{table}
In this section, we present the datasets and models used in experiments, and summarize the statistical data in \cref{tab:dataset_summary}.
\subsection{Natural Language Understanding}
The GLUE Benchmark \citep{wang2019glue} comprises a diverse array of tasks that are widely employed for evaluation in natural language understanding.
It encompasses two single-sentence classification tasks, three tasks assessing similarity and paraphrasing, and four tasks focusing on natural language inference.
Specifically, it includes MNLI (MultiNLI, \citet{williams2017broad}), SST-2 (Stanford Sentiment Treebank, \citet{socher2013recursive}), MRPC (Microsoft Research Paraphrase Corpus, \citet{dolan2005automatically}), CoLA (Corpus of Linguistic Acceptability, \citet{warstadt2019neural}), QNLI (Question NLI, \citet{rajpurkar2018know}), QQP (Quora Question Pairs, \citet{wang2017bilateral}), RTE (Recognizing Textual Entailment, \citet{dagan2005pascal}), and STS-B (Semantic Textual Similarity Benchmark, \citet{cer2017semeval}).
We summarize the statistical data for all datasets within the GLUE Benchmark in \cref{tab:dataset_summary}.
Following existing practices, the development set is used in GLUE as the test data since the actual test set is not publicly available.
We report the overall (matched and mismatched) accuracy for MNLI, Matthew's correlation for CoLA, Pearson correlation for STS-B, and accuracy for the other tasks.

The Reuters-21578 \citep{padmanabhan2016topic} dataset is one of the most widely used data collections for text categorization research. It was collected from the Reuters financial newswire service in 1987 and is used for text classification and natural language processing tasks.
Three splits are available: ModApte, ModHayes, and ModLewis.
These documents cover various topics, such as politics, economics, and sports.
F1 score is used as the evaluation metric across all three splits.

\subsection{Natural Language Generation}
In our experiments on natural language generation, we use the E2E \citep{novikova2017e2e} dataset, which was initially introduced as a dataset for training end-to-end, data-driven natural language generation systems.
Multiple references can be associated with each source table used as input. Each sample input $(x, y)$ consists of a series of slot-value pairs accompanied by an associated natural language reference text.
The E2E dataset comprises approximately $42$k training examples, $4,600$ validation examples, and $4,600$ test examples from the restaurant domain.

We utilize the following five evaluation metrics: BLEU \citep{papineni2002bleu}, NIST \citep{lin2004automatic}, METEOR \citep{banerjee2005meteor}, ROUGE-L \citep{lin2004rouge}, and CIDEr \citep{vedantam2015cider}.

\subsection{Token Classification}
For token classification, we fine-tune the RoBERTa-base and RoBERTa-large models on the BioNLP dataset \citep{collier2004introduction} and the CoNLL2003 dataset \citep{sang2003introduction}.
BioNLP \citep{collier2004introduction} is a Named Entity Recognition dataset that contains biological entities such as DNA, RNA, and protein.
It is essentially a token classification task where we want to classify each entity in the sequence.
CoNLL-2003 \citep{sang2003introduction} focuses on language-independent named entity recognition.
It concentrates on four types of named entities: persons, locations, organizations, and miscellaneous entities that do not belong to the previous three groups.
Accuracy, precision, recall, and F1 score are used as evaluation metrics.

\subsection{Biomedical Experiments}
The ESM (Evolutionary Scale Modeling, \citet{rives2021biological}) model is a transformer-based protein language model designed for protein sequence analysis, leveraging the transformer architecture to capture evolutionary patterns.
We fine-tune the ESM model using the Protein Aligner checkpoint \citep{zhang2024proteinaligner} on two classification tasks—thermostability prediction (\citet{chenhotprotein}, 936 training samples) and blood-brain barrier peptide prediction (BBP, \citet{dai2021bbppred}, 200 training samples)—and one regression task, minimum inhibitory concentration prediction (MIC, \citet{ledesma2023engineered}, 3,695 training samples).
Notably, protein analysis datasets are typically much smaller than those in NLP, in which case the large pre-trained models are prone to overfitting, even when using PEFT methods.
The trainable parameters (on the order of millions) are significantly overparameterized compared to the available samples (thousands or even hundreds), highlighting the need for our overfitting-resilient counterpart.

\section{Experimental Settings}
\label{sec:appendixexperimentalsettings}
In this section, we provide detailed experimental settings.
We maintain consistent configurations across experiments, including LoRA rank, LoRA $\alpha$, batch size, maximum sequence length, and optimizer, to ensure a fair comparison.
For results other than \cref{tab:additional}, we do not include the bias term in PEFT linear layers.
The hyperparameter tuning for our method is straightforward and convenient.

\subsection{RoBERTa}
We summarize the experimental settings for the GLUE benchmark (\cref{tab:nlu}) and for the Reuters21578 dataset and token classification (\cref{tab:combined_results}) tasks in \cref{tab:all_hyperparameters}.

\begin{table*}[htbp!]
    \centering
    \caption{The hyperparameters used for RoBERTa on the GLUE benchmark \citep{wang2019glue}, Reuters21578 dataset \citep{padmanabhan2016topic}, BioNLP dataset \citep{collier2004introduction}, and CoNLL2003 dataset \citep{sang2003introduction}.}
    \vspace{0.2cm}
    \label{tab:all_hyperparameters}
    \setlength{\tabcolsep}{2.5pt} %
    \resizebox{\textwidth}{!}{%
    \begin{tabular}{@{}llccccccccccccc@{}}
        \toprule
        & Settings & MNLI & SST-2 & MRPC & CoLA & QNLI & QQP & RTE & STS-B & ModApte & ModHayes & ModLewis & BioNLP & CoNLL2003\\
        \midrule
        & Optimizer & \multicolumn{13}{c}{AdamW} \\
        & Warmup Ratio & \multicolumn{13}{c}{$0.06$} \\
        & Scheduler & \multicolumn{13}{c}{Linear} \\
        & LoRA rank & \multicolumn{13}{c}{rank = $4$} \\
        & LoRA $\alpha$ & \multicolumn{13}{c}{$8$} \\
        \midrule
        \multirow{8}{*}{\rotatebox{90}{RoBERTa-base}} & Total batch size & \multicolumn{13}{c}{$32$} \\
        & Global steps & $20$k & $12$k & $25$k & $20$k & $15$k & $20$k & $15$k & $12$k & $20$k & $20$k & $20$k & $12$k & $12$k\\
        & Lower learning rate & 5e-5 & 1e-5 & 2e-5 & 5e-5 & 2e-5 & 5e-5 & 1e-5 & 1e-5 & 3e-5 & 3e-5 & 3e-5 & 1e-5 & 2e-5\\
        & Upper learning rate & 5e-5 & 1e-5 & 2e-5 & 5e-5 & 2e-5 & 5e-5 & 1e-5 & 1e-5 & 3e-5 & 3e-5 & 3e-5 & 1e-5 & 2e-5\\
        & Lower weight decay & $0.1$ & $0.1$ & $0.1$ & $0.1$ & $0.1$ & $0.1$ & $0.1$ & $0.1$ & $0.1$ & $0.1$ & $0.1$ & $0.1$ & $0.2$\\
        & Upper weight decay & $0.1$ & $0.1$ & $0.1$ & $0.1$ & $0$ & $0.1$ & $0.1$ & $0.01$ & $0.1$ & $0.1$ & $0.1$ & $0.1$ & $0.1$\\
        & Max Seq Length & \multicolumn{13}{c}{$512$}\\
        & Regularization Coefficient & 1e-5 & 1e-5 & 1e-5 & 1e-5 & 1e-5 & 1e-5 & 1e-5 & 1e-5 & 0 & 1e-5 & 0 & 1e-5 & 0\\
        \midrule
        \multirow{8}{*}{\rotatebox{90}{RoBERTa-large}} & Total batch size & \multicolumn{13}{c}{$32$} \\
        & Global steps & $50$k & $20$k & $30$k & $20$k & $60$k & $40$k & $15$k & $10$k & $20$k & $20$k & $20$k & $12$k & $15$k\\
        & Lower learning rate & 1e-5 & 1e-5 & 1e-5 & 1e-5 & 1e-5 & 1e-5 & 1e-5 & 1e-5 & 1e-5 & 1e-5 & 1e-5 & 2e-5 & 1e-5\\
        & Upper learning rate & 1e-5 & 1e-5 & 1e-5 & 1e-5 & 1e-5 & 1e-5 & 1e-5 & 1e-5 & 1e-5 & 1e-5 & 1e-5 & 2e-5 & 1e-5\\
        & Lower weight decay & $0.5$ & $0.5$ & $0$ & $0.2$ & $0.5$ & $0.5$ & $0.5$ & $0.5$ & $0.2$ & $0.1$ & $0.2$ & $0.02$ & $0.1$\\
        & Upper weight decay & $0.5$ & $0.05$ & $0$ & $0.2$ & $0.5$ & $0.5$ & $0.1$ & $0.5$ & $0.1$ & $0.1$ & $0.1$ & $0.02$ & $0.1$\\
        & Max Seq Length & \multicolumn{13}{c}{$128$}\\
        & Regularization Coefficient & 0 & 0 & 1e-5 & 1e-5 & 0 & 1e-5 & 0 & 1e-5 & 0 & 1e-5 & 0 & 0 & 1e-5\\
        \bottomrule
    \end{tabular}%
    }
\end{table*}

\subsection{GPT-2}
We summarize the experimental settings for the GPT-2 experiments (\cref{tab:nlg}) in \cref{tab:gpt2robertahyperparameters}.
The experimental configuration, particularly during the inference stage, follows the approach described by \citet{hulora}.
\begin{table}[h]
    \centering
    \caption{The hyperparameters we used for GPT-2 on the E2E NLG benchmark \citep{novikova2017e2e}.}
    \vspace{0.2cm}
    \label{tab:gpt2robertahyperparameters}
    \begin{tabular}{lc}
        \toprule
        Settings & Training \\
        \midrule
        Optimizer & AdamW \\
        Warmup Ratio & $0.06$ \\
        Scheduler & Linear \\
        LoRA rank & rank\textsubscript{a} = rank\textsubscript{u} = $4$ \\
        LoRA \(\alpha\) & $32$ \\
        Label Smooth & $0.1$ \\
        Lower learning rate & 1e-3\\
        Upper learning rate & 1e-4\\
        Lower weight decay & $1$ \\
        Upper weight decay & $1$ \\
        Max Seq Length & $512$\\
        Regularization Coefficient & 1e-5\\
        \midrule
        Settings & Inference \\
        \midrule
        Beam Size & $10$ \\
        Length Penalty & $0.9$ \\
        no repeat ngram size & $4$ \\
        \bottomrule
    \end{tabular}
\end{table}

\section{Baselines in Experiments}
\label{sec:baselines}

Here, we provide a brief introduction to compare baselines in all our experiments.
\begin{itemize}
    \item \textbf{Full Fine-Tuning (FT):} The entire model is fine-tuned, with updates to all parameters.
    \item \textbf{Adapter Tuning~\citep{houlsby2019parameter,lin2020exploring,ruckle2020adapterdrop,pfeiffer2020adapterfusion}:} Methods that introduce adapter layers between the self-attention and MLP modules for parameter-efficient tuning.
    \item \textbf{LoRA~\citep{hu2022lora}:} A method that estimates weight updates via low-rank matrices.
    \item \textbf{AdaLoRA~\citep{zhang2023adaptive}:} An extension of LoRA that dynamically reallocates the parameter budget based on importance scores.
    \item \textbf{DoRA~\citep{liu2024dora}:} Decomposes pretrained weights into magnitude and direction, using LoRA for efficient directional updates.
    \item \textbf{VeRA~\citep{kopiczko2023vera}:} Employs a single pair of low-rank matrices across all layers to reduce trainable parameters.
    \item \textbf{FourierFT~\citep{gao2024parameter}:} Fine-tunes models by learning a subset of spectral coefficients in the Fourier domain.
    \item \textbf{AFLoRA~\citep{liu2024aflora}:} Freezes low-rank adaptation parameters using a learned score, reducing trainable parameters while maintaining performance.
    \item \textbf{LaMDA~\citep{azizi2024lamda}:} Fine-tunes large models via spectrally decomposed low-dimensional adaptation.
    \item \textbf{SSH~\citep{shen2025ssh}:} Fine-tunes large models after transforming weight matrices with the discrete Hartley transformation (DHT).
    \item \textbf{MaCP~\citep{shen2025macp}:} Fine-tunes large models by projecting the low-rank adaptation weight change into the discrete cosine space.
\end{itemize}

\section{Weight Decomposition Analysis}
\label{sec:decomposition}
We provide a brief review of the weight decomposition analysis proposed in \citet{liu2024dora}. 
Define the weight decomposition of a weight matrix $\mathbf{W} \in \mathbb{R}^{d \times k}$ (e.g., query matrix in an attention layer) as $\mathbf{W} = \mathbf{m}\frac{\mathbf{V}}{\|\mathbf{V}\|_c} = \|\mathbf{W}\|_c\frac{\mathbf{W}}{\|\mathbf{W}\|_c}$,
where $\mathbf{m} \in \mathbb{R}^{1 \times k}$ is the magnitude vector, and $\mathbf{V} \in \mathbb{R}^{d \times k}$ is the directional matrix, with $\|\cdot\|_c$ representing the vector-wise norm of a matrix across each column. This decomposition ensures that each column of $\mathbf{V} / \|\mathbf{V}\|_c$ remains a unit vector, and the corresponding scalar in $\mathbf{m}$ defines the magnitude of each vector.
\citet{liu2024dora} examine the magnitude and directional variations between $\mathbf{W_0}$ and $\mathbf{W}_{\mathrm{FT}}$, defined as 
$\Delta \mathbf{M}_{\mathrm{FT}}^t = \frac{\sum_{n=1}^{k}|\mathbf{m}_{\mathrm{FT}}^{n,t} - \mathbf{m}_{0}^{n}|}{k}$ and
$\Delta \mathbf{D}_{\mathrm{FT}}^t = \frac{\sum_{n=1}^k(1 - \cos(\mathbf{V}_{\mathrm{FT}}^{n,t}, \mathbf{W_0}^n))}{k}$.
Here, $\Delta \mathbf{M}_{\mathrm{FT}}^{t}$ and $\Delta \mathbf{D}_{\mathrm{FT}}^{t}$ represent the magnitude and direction differences between $W_{0}$ and $W_{\mathrm{FT}}$ at the $t$-th training step, respectively, with $\cos(\cdot, \cdot)$ denoting cosine similarity. 
$m_{\mathrm{FT}}^{n,t}$ and $m_0^n$ are the $n^{\text{th}}$ scalars in their respective magnitude vectors, while $\mathbf{V}_{\mathrm{FT}}^{n,t}$ and $\mathbf{W_0}^n$ are the $n^{\text{th}}$ columns in $\mathbf{V}_{\mathrm{FT}}^{t}$ and $\mathbf{W_0}$.
Intuitively, a consistent positive slope trend across all the intermediate steps implies a difficulty in concurrent learning of both magnitude and direction, suggesting that slight directional changes are challenging to execute alongside more significant magnitude alterations.
In contrast, a relatively negative slope signifies a more varied learning pattern, with a more pronounced negative correlation indicating a larger learning capacity.

\begin{figure}[htbp]
    \centering
    \subfigure[FT($k=-49.279$)]{\includegraphics[width=0.24\textwidth]{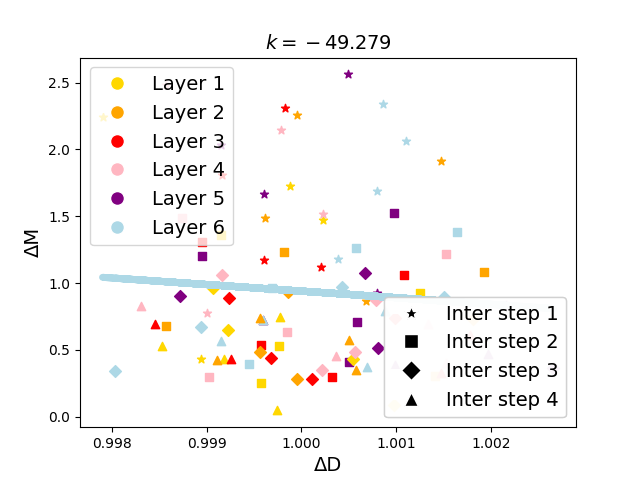}}
    \subfigure[LoRA($k=2.503$)]{\includegraphics[width=0.24\textwidth]{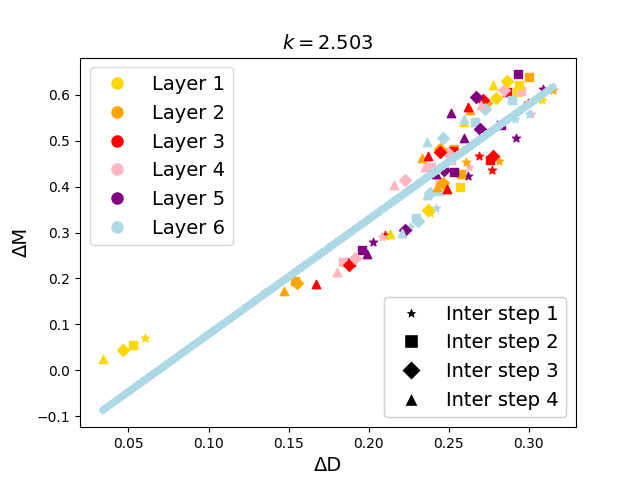}}
    \subfigure[DoRA($k=-5.485$)]{\includegraphics[width=0.24\textwidth]{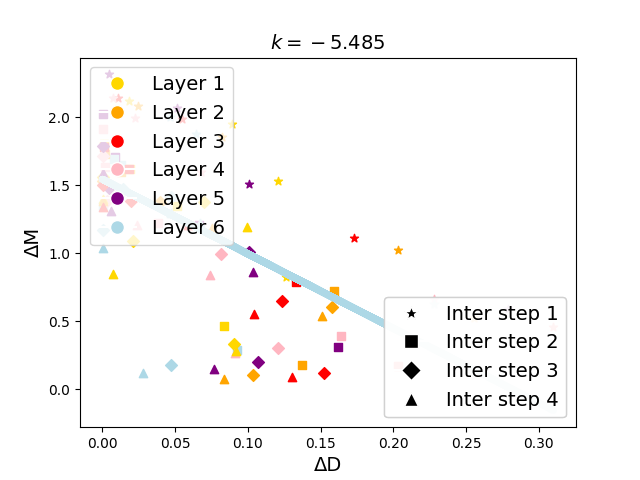}}
    \subfigure[BiDoRA($k=-10.5$)]{\includegraphics[width=0.24\textwidth]{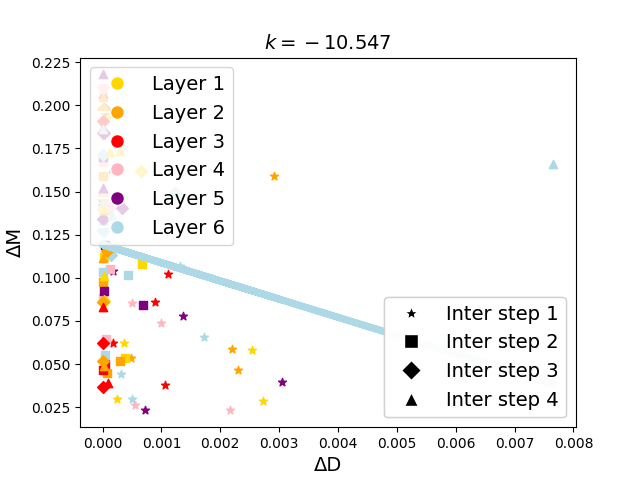}}
    \caption{\textbf{Magnitude and direction updates} for (a) FT, (b) LoRA, (c) DoRA, and (d) BiDoRA of the value matrices across different layers and intermediate steps after fine-tuning the GPT2 model on the E2E dataset \citep{novikova2017e2e}.
Different markers represent matrices from different training steps, while different colors indicate matrices from each layer.
The values of negative correlation are shown at the top, denoted by $k$.
}
    \label{fig:wdavalue}
\end{figure}

Complementary to \cref{fig:wdaquery} in the main paper on the query matrix, we provide additional results of weight decomposition analysis in \cref{fig:wdavalue} on the value matrix to complement the findings in \cref{sec:wda}.
We can draw two key observations from \cref{fig:wdavalue}: 1) Consistent with the results in \citet{liu2024dora}, both FT and DoRA exhibit negative correlation values of $-49.279$ and $-5.485$, respectively, while LoRA shows a positive correlation with a value of $2.503$. 2) BiDoRA achieves a negative correlation value of $-10.547$, indicating closer alignment with FT compared to DoRA.
The analysis of how BiDoRA achieves this improvement is similar to that discussed in \cref{sec:wda}.

\section{Training Cost}
\label{sec:cost}
\begin{table}[htbp]
    \centering
        \caption{Average training time cost on the GLUE benchmark \citep{wang2019glue}.}
    \vspace{0.2cm}
    \begin{tabular}{c|c|c|c}
        \toprule
        Method            & LoRA & DoRA & BiDoRA \\
        \midrule
        Per-step cost     & $\times1$      & $\times1.36$           & $\times3.54$   \\
        Total steps       & $27.45$k       & $27.45$k               & $17.37$k   \\
        Total time        & $\times1$      & $\times1.36$           & $\times2.24$   \\
        \bottomrule
    \end{tabular}
    \label{tab:trainingcost}
\end{table}
\cref{tab:trainingcost} compares the training efficiency of LoRA, DoRA, and BiDoRA on the GLUE benchmark using the RoBERTa-base model.
The table details the total training steps required for convergence and the per-step computational cost, which is normalized relative to LoRA for reference.
For a fair comparison, all methods were benchmarked on a single NVIDIA A100 GPU.
The results show that BiDoRA converges in fewer steps than LoRA and DoRA, while the per-step cost for BiDoRA is modestly higher, as its BLO process requires iterative updates between the two levels and the computation of hypergradients.
The total training time for BiDoRA is approximately $1.64$ times that of DoRA, a training cost that remains comparable to the baselines.
Given BiDoRA's superior performance across various tasks, we argue that this slight increase in computational cost is an acceptable trade-off, underscoring our method's practicality.

\section{The Role of Hyperparameter}
The hyperparameter tuning for BiDoRA is simple, convenient, and straightforward.
We further conducted experiments regarding the dataset partition of $\trainset$ and $\valset$ to provide insights into its role in BiDoRA.
The dataset partition helps maintain the balance of inner/outer optimization by assigning different portions of data.
The direction component has more trainable parameters, so it is reasonable to use more data for training the lower level while using the remaining data for training magnitudes.
As shown in \cref{tab:split}, we varied the inner-level dataset $\trainset$ partition from $0.6$ to $1.0$ with $0.1$ intervals and experimented with RoBERTa-base on three splits of the Reuters21578 dataset to examine its influence.

The results indicate that both extreme cases are negative to the overall performance.
When the inner partition is too small ($\leq 0.6$), directions are not well-trained, and when the inner partition is $1.0$, magnitudes are not trained at all, leading to a significant performance drop.
These findings demonstrate that BLO is effective in the sense that both levels are necessary for enhancing performance.
Although tuning the partition ratio may further improve overall performance, we maintain a consistent data partition of $8:2$ in all the experiments for simplicity.
A fixed configuration of data partition already consistently yields superior performance of BiDoRA, demonstrating that our method is robust to this hyperparameter within a certain range.

\begin{table}
    \centering
    \caption{Experiment results on different data partitions of BiDoRA.}
    \vspace{0.2cm}
    \begin{tabular}{c|ccc}\toprule
        Partition & ModApte & ModHayes & ModLewis \\ \midrule
        $0.6$ & $85.32$ & $79.76$ & $77.69$\\
        $0.7$ & $85.32$ & $\textbf{80.01}$ & $\textbf{77.74}$ \\
        $0.8$ & $\textbf{85.34}$ & $79.93$ & $77.63$ \\
        $0.9$ & $85.27$ & $79.85$ & $77.64$\\
        $1.0$ & $85.23$ & $79.59$ & $77.42$\\ \bottomrule
    \end{tabular}
    \label{tab:split}
\end{table}

\section{Comparison with Other General Methods for Addressing Overfitting}
There are some common experimental settings that may be used to reduce overfitting.
For DoRA, two promising methods are increasing weight decay and adopting a more aggressive dropout rate.
We conducted experiments on these two methods separately.
We kept hyperparameters that have been well-tuned in DoRA and can achieve optimal results while only tuning the weight decay value.
Similarly, we tune the dropout rate of DoRA while keeping the weight decay value to be optimized.
We conducted experiments on RoBERTa-base on three datasets.
The results are presented in \cref{tab:other}.

We can draw the observation that neither of these approaches effectively addresses overfitting issues or enhances the model's generalization ability.
On the other hand, BiDoRA exploits the specific magnitude-direction structure of DoRA and the strategy of training the two distinct components on separate splits of the dataset.
An advantage of our methodology is that it can be easily combined with other general-purpose overfitting-alleviating strategies since BiDoRA does not modify the original DoRA architecture.

\begin{table}
    \centering
    \caption{Experiment results on different weight decay values and different dropout rates of DoRA.}
    \vspace{0.2cm}
    \begin{tabular}{c|ccc}\toprule
        Method & CoLA & MRPC & RTE \\ \midrule
        DoRA (weight decay = 0)    & 59.3 & 88.7 & 72.9  \\
        DoRA (weight decay = 0.05) & 60.1 & 89.2 & 73.3\\
        DoRA (weight decay = 0.1)  & 60.5 & 89.2 & 73.2 \\
        DoRA (weight decay = 0.2)  & 60.3 & 89.0 & 73.2\\ \midrule
        DoRA (dropout rate = 0)   & 59.2 & 89.2 & 72.9\\
        DoRA (dropout rate = 0.1) & 60.2 & 88.9 & 71.4\\
        DoRA (dropout rate = 0.2) & 55.1 & 87.8& 64.2\\ \midrule
        BiDoRA & \textbf{61.3} & \textbf{89.4} & \textbf{76.0} \\ \bottomrule
    \end{tabular}
    \label{tab:other}
\end{table}

\section{Additional Experiments}
\subsection{Experiments on Natural Language Generation Tasks}
\label{sec:nlg}
In this section, we evaluate BiDoRA's performance on the NLG task.
\cref{tab:nlg} presents the results of fine-tuning a GPT-2 model on the E2E dataset with baseline PEFT methods and BiDoRA. The results show that BiDoRA achieves the best performance across all five evaluation metrics, demonstrating the superiority of BiDoRA in fine-tuning pre-trained models for NLG tasks.

\begin{table}[htbp]
    \centering
    \caption{Performance of BiDoRA and baseline methods for fine-tuning GPT2-medium on the E2E dataset \citep{novikova2017e2e}. A higher value is better for all metrics. The best results are shown in \textbf{bold}.}
    \vspace{0.2cm}
    \begin{tabular}{c|c|ccccc} \toprule
        Method & \#Params & BLEU & NIST  & MET & ROUGE-L & CIDEr        \\ \midrule
        FT &$354.9$M & $68.0$ &$8.61$& $46.1$& $69.0$& $2.38$ \\ \midrule
        Adapter &$11.1$M & $67.0$ &$8.50$& $45.2$& $66.9$& $2.31$ \\
        LoRA   & $0.39$M& $67.1$ &$8.54$&$45.7$&$68.0$&$2.33$  \\
        DoRA   & $0.39$M& $67.0$& $8.48$&$ 45.4$ &$70.1$ &$2.33$\\
        BiDoRA & $0.39$M&$\textbf{69.0}$ &$\textbf{8.72}$ &$\textbf{46.2}$ &$\textbf{70.9}$ &$\textbf{2.44}$ \\ \bottomrule
    \end{tabular}
    \vspace{-0.5cm}
    \label{tab:nlg}
\end{table}

\subsection{Experiments on Token Classification}
\label{sec:token}
The effectiveness of BiDoRA can also be observed in \cref{tab:combined_results}, which reports the results of token classification tasks.
Unlike the NLU tasks discussed in the previous section, which involve classifying entire sentences and focusing on capturing global semantics, token classification requires classifying each token within a sentence, highlighting the importance of capturing local context.
On the BioNLP dataset, BiDoRA consistently outperforms baseline methods by a large margin in terms of F1 score.
On the CoNLL2003 dataset, BiDoRA either outperforms or matches all baseline methods across all metrics.
Consistent with our previous findings, BiDoRA effectively fine-tunes pre-trained models for token classification tasks.

\subsection{More Experiments on Natural Language Understanding Tasks} 
\cref{tab:combined_results} presents the results of fine-tuning RoBERTa models on the Reuters21578 datasets, a text classification task, where BiDoRA outperforms all baseline methods by an even larger margin. 
Notably, BiDoRA achieves performance comparable to or even better than full fine-tuning, providing further evidence of its superiority.

\begin{table*}[htbp!]
    \centering
    \caption{
    RoBERTa\textsubscript{base/large} (R\textsubscript{b/l}) with different fine-tuning methods on the Reuters21578 \citep{padmanabhan2016topic}, BioNLP \citep{collier2004introduction}, and CoNLL2003 \citep{sang2003introduction} benchmarks.
    A higher value is better for all metrics. The best results are shown in \textbf{bold}.
    }
    \label{tab:combined_results}
    \setlength{\tabcolsep}{1pt}
    \vspace{0.2cm}
    \small
    \begin{tabular}{c|c|ccc|cccc|cccc}
        \toprule
        & & \multicolumn{3}{c|}{Reuters21578} & \multicolumn{4}{c|}{BioNLP} & \multicolumn{4}{c}{CoNLL2003} \\
        \cmidrule(lr){3-5} \cmidrule(lr){6-9} \cmidrule(lr){10-13}
        Method & \#Params& ModApte & ModHayes & ModLewis & Accuracy & Precision & Recall & F1 & Accuracy & Precision & Recall & F1 \\
        \midrule
        R\textsubscript{b}(FT)     & $125.0$M& $85.4$&$77.6$&$77.1$ & $93.9$&$69.0$&$78.9$&$73.6$&$99.3$&$95.7$&$96.3$&$96.0$ \\
        \midrule
        R\textsubscript{b}(Adapter)& $0.9$M& $\textbf{85.3}$&$77.5$&$76.8$ & $93.9$&$69.1$&$78.8$&$73.7$&$\textbf{99.3}$&$95.7$&$96.4$&$96.0$ \\
        R\textsubscript{b}(LoRA)   & $0.15$M& $84.7$ & $74.3$ & $74.7$ & $93.9$ &  $69.0$ & $78.8$ & $73.6$  & $\textbf{99.3}$&$95.4$&$96.3$  &  $95.8$  \\
        R\textsubscript{b}(DoRA)   & $0.17$M& $84.8$ & $78.2$ & $76.6$ & $\textbf{94.0}$ &  $69.2$ & $\textbf{79.1}$ & $73.8$ & $\textbf{99.3}$ & $95.3$   &  $96.2$  &  $95.8$  \\
        R\textsubscript{b}(BiDoRA) & $0.17$M& $\textbf{85.3}$ & $\textbf{79.9}$ & $\textbf{77.6}$ & $93.9$ & $\textbf{71.2}$ & $78.6$ & $\textbf{74.7}$  & $\textbf{99.3}$& $\textbf{95.9}$&  $\textbf{96.5}$  &  $\textbf{96.2}$\\
        \midrule \midrule
        R\textsubscript{l}(FT)     & $355.0$M& $84.8$&$77.5$& $76.6$ & $94.0$&$69.4$&$79.6$&$74.1$&$99.4$&$96.2$&$97.0$&$96.6$ \\
        \midrule
        R\textsubscript{l}(Adapter)& $0.44$M& $84.8$&$77.9$& $76.7$ & $\textbf{94.0}$&$69.4$&$79.7$&$74.2$&$\textbf{99.4}$&$96.1$&$97.0$&$96.6$\\
        R\textsubscript{l}(LoRA)   & $0.39$M& $84.7$ & $77.7$ & $76.7$ & $93.9$ &  $69.2$ & $79.3$ & $73.9$& $\textbf{99.4}$ & $96.2$   &  $97.0$  &  $96.6$ \\
        R\textsubscript{l}(DoRA)   & $0.39$M& $84.8$ & $77.4$ & $76.7$ & $\textbf{94.0}$ &  $69.4$ & $\textbf{79.7}$ & $74.2$& $\textbf{99.4}$&  $96.2$  &  $\textbf{97.1}$  & $96.6$  \\
        R\textsubscript{l}(BiDoRA) & $0.39$M& $\textbf{84.9}$ & $\textbf{78.9}$ & $\textbf{77.3}$ & $\textbf{94.0}$ &  $\textbf{71.3}$ & $79.3$ & $\textbf{75.1}$  & $\textbf{99.4}$ & $\textbf{96.4}$ &$\textbf{97.1}$& $\textbf{96.7}$ \\
        \bottomrule
    \end{tabular}
\end{table*}

\section{Evidence on Orthogonality of Incremental Matrix}

\begin{figure*}
    \centering
    \subfigure[DoRA-query matrix]{\includegraphics[width=0.32\textwidth]{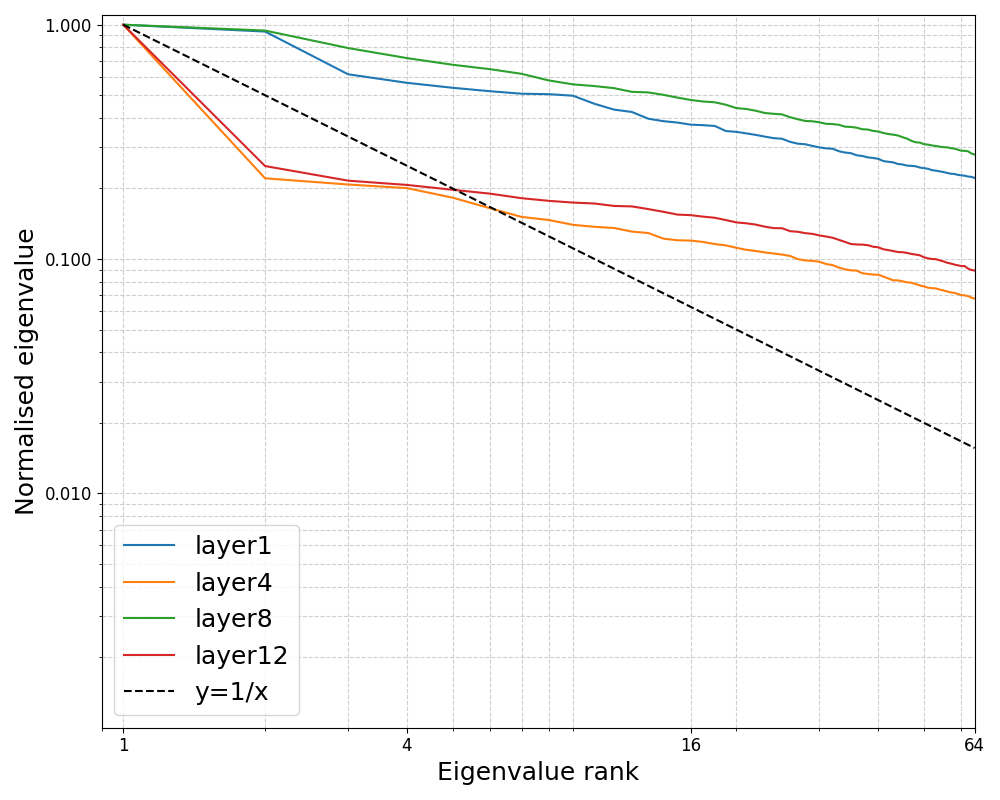}}
    \subfigure[BiDoRA (w/o cst.)-query matrix]{\includegraphics[width=0.32\textwidth]{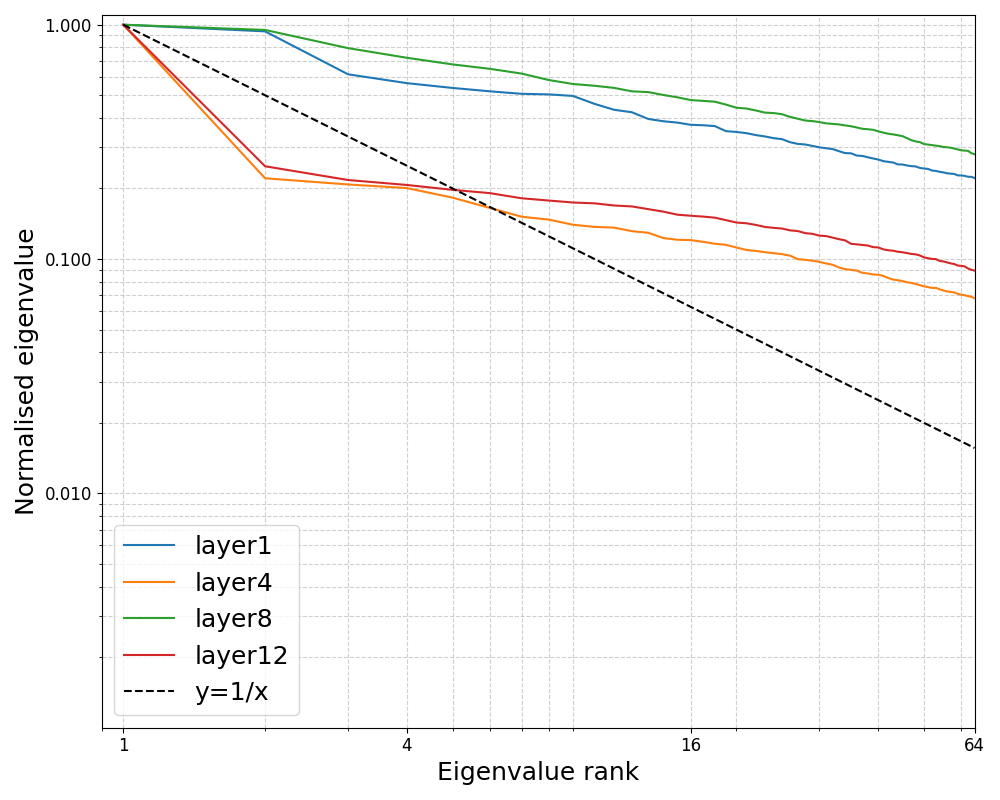}}
    \subfigure[BiDoRA-query matrix]{\includegraphics[width=0.32\textwidth]{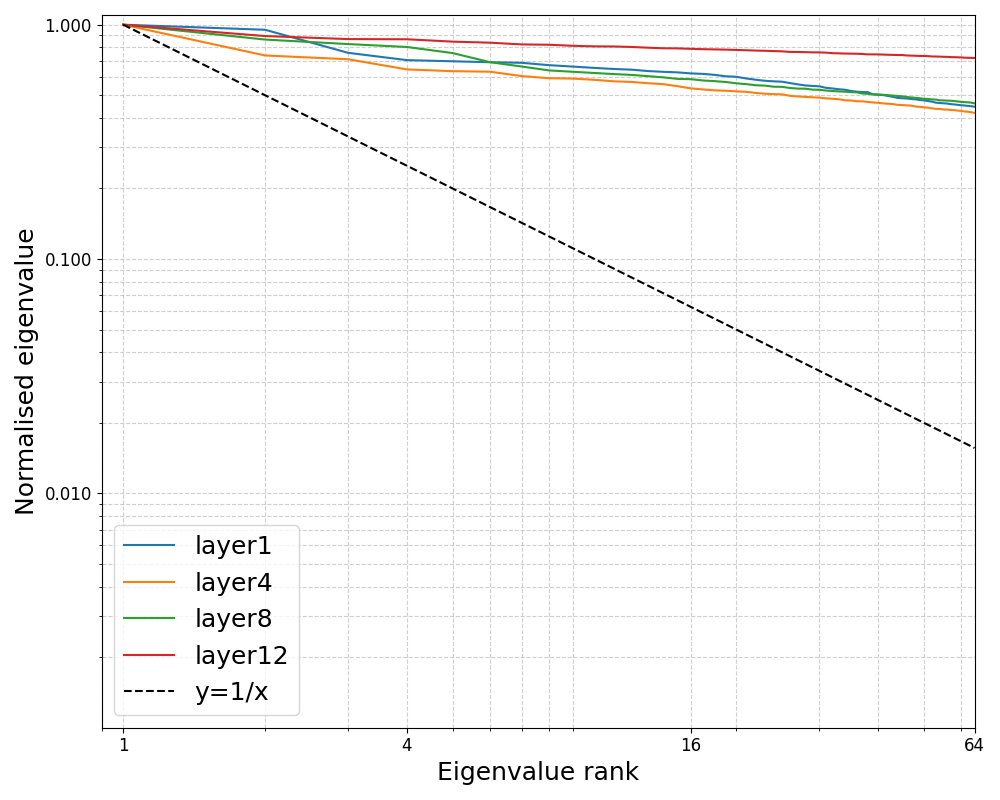}}
    \subfigure[DoRA-value matrix]{\includegraphics[width=0.32\textwidth]{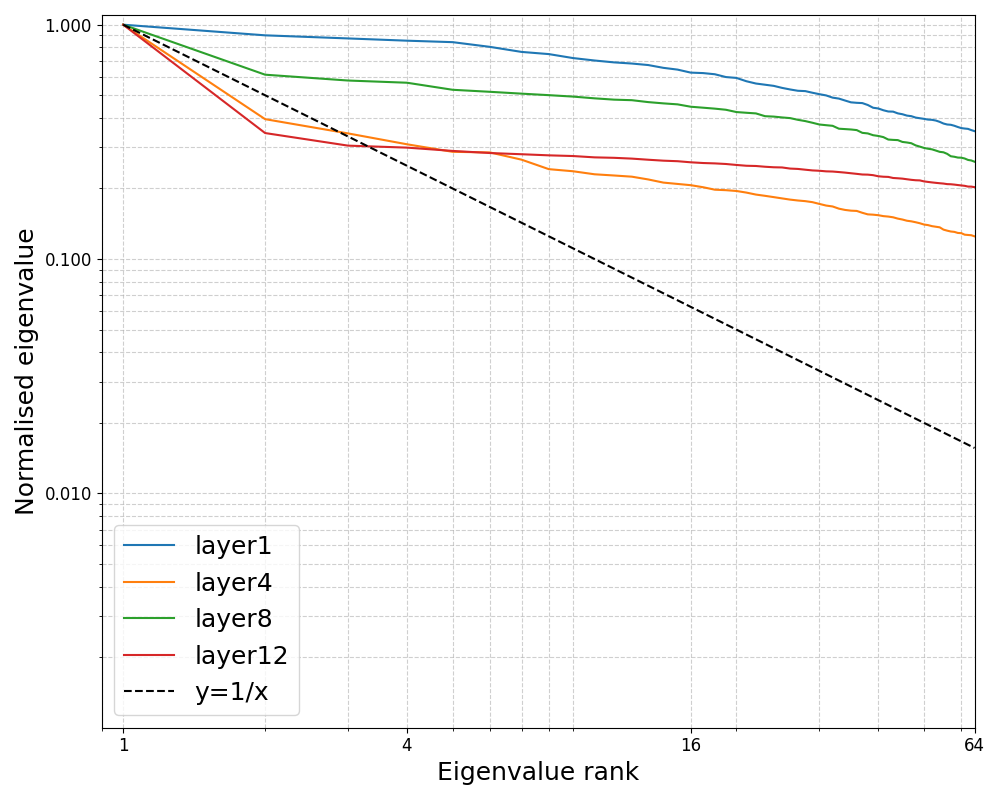}}
    \subfigure[BiDoRA (w/o cst.)-value matrix]{\includegraphics[width=0.32\textwidth]{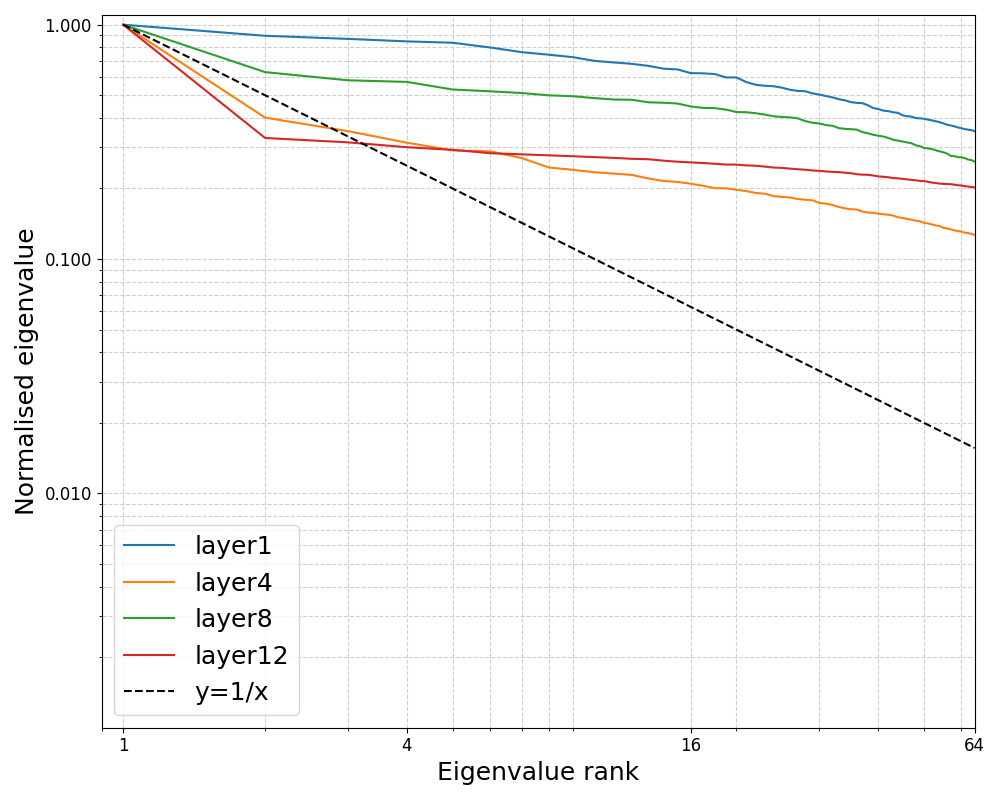}}
    \subfigure[BiDoRA-value matrix]{\includegraphics[width=0.32\textwidth]{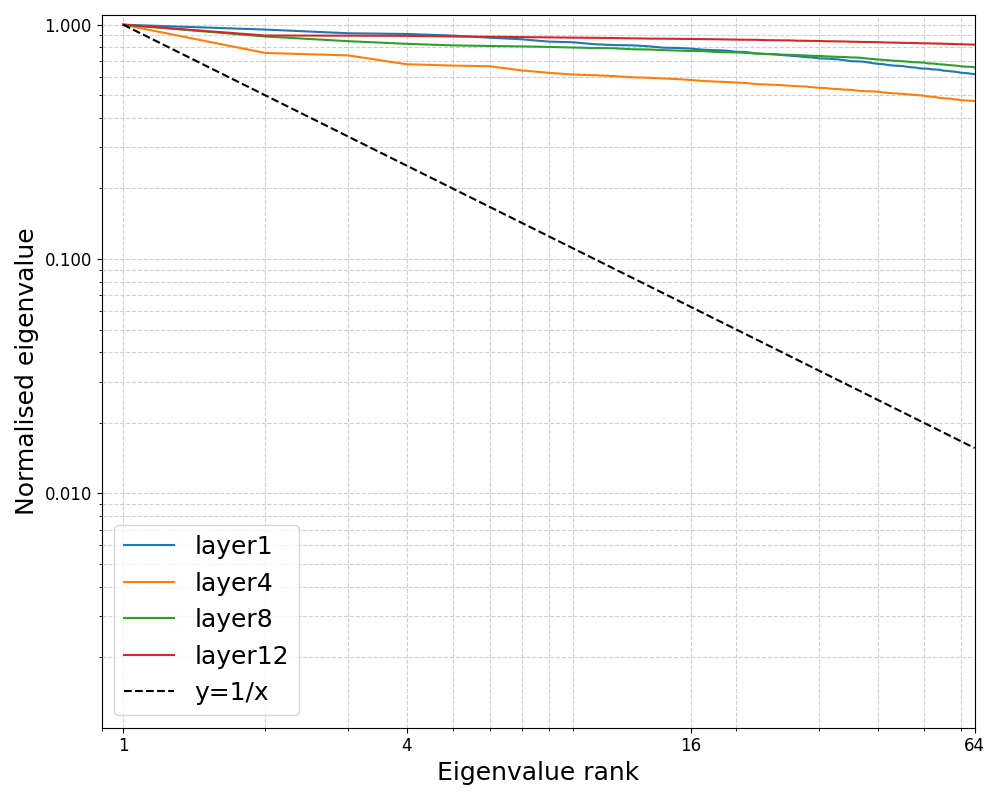}}
    \caption{Eigenspectra of the direction matrix for query (top) and value (bottom) matrices across different layers. The figure compares three fine-tuning methods: BiDoRA, BiDoRA without orthogonal regularization (w/o cst.), and DoRA. Both axes are on a log scale, and only the $64$ largest eigenvalues are shown for visualization clarity. Experiments were conducted on the CoLA dataset \citep{warstadt2019neural} with the RoBERTa-base model.}
    \label{fig:or}
\end{figure*}

To verify that the orthogonal regularization (OR) proposed in \cref{sec:or} encourages the columns of the direction matrix to be orthogonal, we visualize the normalized eigenvalues of the matrix in \cref{fig:or}.
The results show that, compared to methods without OR (i.e., DoRA and BiDoRA w/o cst.), BiDoRA with OR produces eigenvalues that are more closely aligned with those of a purely orthogonal matrix, where all eigenvalues would be one. 
This effect holds for both the query and value matrices and verifies the effectiveness of the OR constraint.

\end{document}